%% file: 0_Final_Copy.tex
\documentclass[11pt,letterpaper]{article}
\pdfoutput=1 
\usepackage[top=3cm, bottom=3cm, left=2.5cm, right=2.5cm]{geometry}
\usepackage{amsmath, amssymb, amsthm, mathabx, mathrsfs, bbm}
\usepackage{graphicx, color}
\usepackage{array, multirow}
\usepackage[font=small,labelfont=bf]{caption} 
\usepackage{lipsum}
    \usepackage[T1]{fontenc}
    \usepackage[utf8]{inputenc} 
    \usepackage{lmodern}
\usepackage{amsfonts}
\usepackage{tabularx}
\usepackage{booktabs}
\usepackage{graphicx}
\usepackage{epstopdf}
\usepackage{algorithmic}
\usepackage{mathtools}
\usepackage{tikz}
\usepackage{todonotes}
\usepackage[normalem]{ulem}
\definecolor{darkgreen}{rgb}{0, .5, 0}
\ifpdf
\DeclareGraphicsExtensions{.eps,.pdf,.png,.jpg}
\else
\DeclareGraphicsExtensions{.eps}
\fi

\AtBeginDocument{%
\setlength{\oddsidemargin}{\dimexpr(\paperwidth-\textwidth)/2-1in}%
\setlength{\evensidemargin}{\oddsidemargin}%
\setlength{\topmargin}{%
\dimexpr(\paperheight-\textheight)/2-\headheight-\headsep-1in}%
}

\usepackage[shortlabels]{enumitem}

\usepackage{longtable} 

\usepackage[sort]{natbib}

\setcitestyle{numbers,square,comma}
    \usepackage[pdfauthor={Nils Detering, Luca Galimberti, Anastasis Kratsios, Giulia Livieri, A. Martina Neuman},
                pdftitle={Learning from one graph: transductive learning guarantees via the geometry of small random worlds},
                pdfsubject={Statistical Learning Theory},
                pdfkeywords={Transductive Learning, Graph Neural Networks, Metric Embedding, Random Graphs, Generalization Bounds, Discrete Geometry, Non-Vacuous}]{hyperref}
    \hypersetup{colorlinks=true,
    urlcolor=blue,
    linkcolor=blue,
    citecolor=blue,
    bookmarksdepth=paragraph}
\usepackage[nameinlink]{cleveref}
\crefname{equation}{}{}
\crefname{section}{section}{sections}
\crefname{figure}{figure}{figures}
\crefname{table}{table}{tables}
\crefname{example}{example}{examples}
\crefname{proposition}{proposition}{propositions}
\Crefname{section}{Section}{Sections}
\Crefname{figure}{Figure}{Figures}
\Crefname{table}{Table}{Tables}
\Crefname{definition}{Definition}{Definitions}
\Crefname{theorem}{Theorem}{Theorems}
\Crefname{remark}{Remark}{Remarks}
\Crefname{example}{Example}{Examples}
\Crefname{proposition}{Proposition}{Propositions}
\numberwithin{equation}{section}

\newtheorem{theorem}{Theorem}[section]
\newtheorem{lemma}{Lemma}[section]
\newtheorem{corollary}{Corollary}[section]
\newtheorem{proposition}{Proposition}[section]
\theoremstyle{definition}
\newtheorem{definition}{Definition}[section]
\newtheorem{remark}{Remark}[section]
\newtheorem{assumption}{Assumption}[section]
\newtheorem{example}{Example}[section]

    \newtheoremstyle{theoreminformal}
      {\topsep}   
      {\topsep}   
      {\itshape}  
      {}          
      {\bfseries} 
      {.}         
      {.5em}      
      {}          
    
    \theoremstyle{theoreminformal}
    \newtheorem*{theoreminformal}{Informal theorem}

\newcommand{\R}{\mathbb{R}}
\newcommand{\N}{\mathbb{N}}

\renewcommand{\ge}{	\geq }
\renewcommand{\le}{	\leq }
\renewcommand{\geq}{\geqslant}
\renewcommand{\leq}{\leqslant}

\renewcommand{\tilde}{\widetilde}

\renewcommand{\epsilon}{\varepsilon}

\definecolor{amethyst}{rgb}{0.6, 0.4, 0.8}
\definecolor{huntergreen}{rgb}{0.21, 0.37, 0.23}
\definecolor{lavenderindigo}{rgb}{0.58, 0.34, 0.92}
\definecolor{lustred}{rgb}{0.9, 0.13, 0.13}
\definecolor{mediumpersianblue}{rgb}{0.0, 0.4, 0.65}
\definecolor{mediumseagreen}{rgb}{0.24, 0.7, 0.44}
\definecolor{mountainmeadow}{rgb}{0.19, 0.73, 0.56}
\definecolor{myrtle}{rgb}{0.13, 0.26, 0.12}
\definecolor{msugreen}{rgb}{0.09, 0.27, 0.23}

\usepackage{floatrow}
\newfloatcommand{capbtabbox}{table}[][\FBwidth]

\usepackage{blindtext}

\usepackage{multirow}
\usepackage[final]{changes}
\definechangesauthor[name={Anees Kazi}, color=blue]{AK}

\input{Package/ToC_Manipulation}

\usepackage{subcaption}
\usepackage{graphicx}
\title{Learning from one graph: transductive learning guarantees via the geometry of small random worlds}

\author{\fontsize{12}{12}\selectfont
  Nils Detering\thanks{Heinrich Heine University D\"{u}sseldorf, Mathematics Institute.
  \texttt{nils.detering@hhu.de}} \,\,
  Luca Galimberti\thanks{Kings College London, Department of Mathematics. \texttt{luca.galimberti@kcl.ac.uk}} \,\,
  Anastasis Kratsios\thanks{McMaster University, Department of Mathematics and Statistics. The Vector Institute.  \texttt{kratsioa@mcmaster.ca}} \,\,
  Giulia Livieri\thanks{The London School of Economics and Political Science. \texttt{g.livieri@lse.ac.uk}} \,\,
  A. Martina Neuman\thanks{University of Vienna, Faculty of Mathematics. \texttt{neumana53@univie.at.ac}.}
  \normalsize
}

\newcommand{ \eqdef }{
    \ensuremath{\stackrel{\mbox{\upshape\tiny def.}}{=}}
}

\newcounter{termcounter}
\renewcommand{\thetermcounter}{\Roman{termcounter}}
\crefname{term}{term}{terms}
\creflabelformat{term}{#2\textup{(#1)}#3}

\makeatletter
\def\term{\@ifnextchar[\term@optarg\term@noarg}
\def\term@optarg[#1]#2{%
  \textup{#1}%
  \def\@currentlabel{#1}%
  \def\cref@currentlabel{[][2147483647][]#1}%
  \cref@label[term]{#2}}
\def\term@noarg#1{%
  \refstepcounter{termcounter}%
  \textup{(\thetermcounter)}%
  \cref@label[term]{#1}}
\makeatother

\RenewCommandCopy{\overbrace}{\LaTeXoverbrace}
\RenewCommandCopy{\underbrace}{\LaTeXunderbrace}

\usepackage{xparse}
\NewDocumentCommand{\Lout}{o}{%
  \mathtt{B}_{\text{out}}\IfValueT{#1}{^{#1}}%
}

\usepackage[most]{tcolorbox}
\usepackage{etoolbox} 
\usepackage{xcolor}
\definecolor{faintgray}{RGB}{245,245,245}     
\definecolor{faintborder}{RGB}{230,230,230}   
\definecolor{lightblack}{gray}{0.4}           
\newcounter{question}
\newtcolorbox[auto counter, use counter=question]{question}[1][]{
  enhanced,
  colback=faintgray,
  colframe=faintborder,
  boxrule=0.2pt,
  arc=2mm,
  title=\textcolor{lightblack}{\textbf{Question~\thequestion}},
  fonttitle=\bfseries,
  before upper={\centering\itshape},
  after title={\vspace{0.5ex}},
  boxsep=4pt,
  left=6pt,
  right=6pt,
  top=4pt,
  bottom=4pt,
  #1
}

\input{Package/commenting}

\begin{document}

\footnotetext[1]{\textit{All authors contributed equally and are listed in alphabetical order.}}
\maketitle

\begin{abstract}
Since their introduction by Kipf and Welling in $2017$, a primary use of graph convolutional networks is transductive node classification, where missing labels are inferred within a single observed graph and its feature matrix. Despite the widespread use of the network model, the statistical foundations of transductive learning remain limited, as standard inference frameworks typically rely on multiple independent samples rather than a single graph.
In this work, we address these gaps by developing new concentration-of-measure tools that leverage the geometric regularities of large graphs via low-dimensional metric embeddings. The emergent regularities are captured using a random graph model; however, the methods remain applicable to deterministic graphs once observed. 
We establish two principal learning results. The first concerns arbitrary deterministic $k$-vertex graphs, and the second addresses random graphs that share key geometric properties with an Erd\H{o}s-R\'{e}nyi graph $\mathbf{G}=\mathbf{G}(k,p)$ in the regime $p \in \mathcal{O}((\log (k)/k)^{1/2})$. The first result serves as the basis for and illuminates the second.
We then extend these results to the graph convolutional network setting, where additional challenges arise.
Lastly, our learning guarantees remain informative even with a few labelled nodes $N$ and achieve the optimal nonparametric rate $\mathcal{O}(N^{-1/2})$ as $N$ grows.
\end{abstract}



\doparttoc 
\faketableofcontents 

\part{} 
\vspace{-3em}
\section{Introduction}
\label{s:Intro}

Graph convolutional networks (GCNs) \citep{kipf2017semi} have rapidly become indispensable in artificial intelligence (AI), powering applications from fake news detection\citep{phan2023fake} and prediction of protein-protein interaction~\citep{jha2022prediction} to early diagnosis of cognitive disorders~\citep{kim2023personalized} and climate modeling~\citep{graphcast2023}. Beyond these, they enable route planning~\citep{wang2024machining}, content recommendation~\citep{zheng2021dgcn}, and personalized online marketplaces~\citep{virinchi2022recommending}. 
Crucially, GCNs can exploit complex graph-based information and the relational structure of data in ways that classical models, such as multi-layer perceptrons (MLPs), cannot, often achieving superior performance on tasks where graph connectivity is central~\citep{bronstein2021geometric,d2024approximation}.
Applications of GCNs can be broadly divided into inductive learning (IL) and transductive learning (TL) \citep{vapnik1982estimation}. In IL, the model learns from multiple, often independent, graph-feature-label triples to make predictions on new, unseen graphs. TL, by contrast, presents a fundamentally different challenge: the learner observes only a single realization of a (possibly random) graph and its node features, along with a subset of node labels, and must infer the remaining labels. Classic TL tasks on graphs include link prediction~\citep{zhang2018link}, such as determining whether two individuals are connected in a single social network snapshot, and node classification~\citep{kipf2017semi}, such as assigning research fields to papers in a partially labeled citation network. Beyond these canonical settings, TL phenomena routinely appear when employing random sub-sampling strategies~\citep{sripathmanathan2023on,jayawant2022practical, neuman2025theoretical} to scale GCNs to massive real-world graphs~\citep{noiseG_polut,oono2020optimization,adam2023zero}, highlighting the broad practical relevance and subtle challenges of TL.

Acquiring labeled data remains a daunting and persistent obstacle in many real-world applications. Human annotation is not only costly and time-consuming, but in numerous scenarios, labels are simply unavailable. 
Furthermore, users often lack access to the multiple independent graph-feature-label triples required for standard IL. 
Instead, they typically work with a single realization of a (possibly random) graph and its node features, with labels available for only a subset of nodes -- i.e. sub-sampling scenario that exemplifies transductive learning. 
The availability of only a \textit{single} sample of the graph and node feature matrix contrasts with the standard statistical setting that underpins IL, which relies on multiple independent samples for inference (the law of large numbers or the central limit theorem). Consequently, TL problems are challenging to study in the absence of such tools, and the corresponding statistical literature remains limited -- often focusing on stylized models~\citep{tang2023information,shi2024homophily} or relying on opaque variants of classical statistical objects, e.g. transductive Rademacher complexities~\citep{el2009transductive,yang2023sharp}. 
By comparison, IL guarantees for GCNs, for example, benefit from a wealth of classical statistical tools, giving rise to a rich and well-developed theoretical framework~\citep{scarselli2018vapnik,garg2020generalization,liao2020pac,levie2023graphon,maskey2025generalization,brilliantov2024compositional}.

It is apparent that establishing robust TL guarantees for GCNs is of paramount importance. In this work, we advance the field by introducing novel geometric tools that expand the statistician’s toolbox, focusing on concentration-of-measure techniques that exploit the emergent geometry of large, dense random graphs via innovative low-dimensional metric embedding arguments. Our transductive learning guarantees are both efficient and powerful: they remain effective when the number of labeled nodes $N$ is small, and they attain the optimal non-parametric rate of $\mathcal{O}(N^{-1/2})$ when $N$ is large, highlighting the robustness of our approach across all regimes.

\subsection{Contributions}

Our main contributions fall into two complementary categories. The first consists of transductive learning guarantees for standard regular graph learners, such as GCNs. Equally important, the second introduces new geometric tools that enrich the statistician's toolbox and may have independent applications beyond GCNs.

\subsection{Main results}
We establish concrete, broadly applicable TL guarantees for suitably regular graph learners (Theorems~\ref{thrm:main_result__deterministic} and~\ref{thrm:Main_Result__random}), treating both the deterministic setting and the \textit{common noise} setting. 
In the former, the guarantees hold for any graph without isolated vertices and with arbitrary node features. (We use ``node'' and ``vertex'' interchangeably.)
In the latter, both the graph and the feature matrix are modeled as single random draws, where the graph has diameter at most $2$ with high probability when the vertex count is large, and the node features are compactly supported.
We present a representative result illustrating the guarantees that hold in the common noise setting for a generalized GNN.

\begin{theoreminformal}[Corollary~\ref{cor:Transductive_GCNNs__random}]
Consider a sufficiently large number of nodes $k$, with labels provided for $N$ sampled nodes. 
Let $\mathbf{X}$ be a $k\times d_{\rm in}$ random feature matrix with bounded i.i.d. entries. 
Let $\mathbf{G}=\mathbf{G}(k,p)$ be an Erd\H{o}s-R\'{e}nyi random graph with $p\in \mathcal{O}((\log (k)/k)^{1/2})$. 
We study the transductive learning task of predicting the remaining labels using models from the generalized GCN class $\mathcal{F}_{\rm GCN}$ (Definition~\ref{defn:GGCN}), trained on the observed pair $(\mathbf{G}, \mathbf{X})$. 
Then for any failure probability $\delta\in (0,1/2)$, the transductive generalization gap holds uniformly over $\mathcal{F}_{{\rm GCN}}$ and is at most
\begin{equation}
    \label{eq:gen_bound}
    C(\theta_{\rm GCN}) \Big(\frac{\min\{\log_2(N), kC'(\theta_{\rm GCN})\}}{N^{1/2}} + \frac{(\log(2/\delta))^{1/2}}{N^{1/2}}
    \Big)
\end{equation}
with probability at least $1-2\delta$. 
Here, $C(\theta_{\rm GCN}), C'(\theta_{\rm GCN})$ denote constants that depend on the network structure $\theta_{\rm GCN}$, encompassing both its parameters and size\footnote{Technically speaking hidden in $\theta_{\rm GCN}$ is a further dependence on $k$. The separation between $k$ and $\theta_{\rm GCN}$ in \eqref{eq:gen_bound} is intended to highlight the additional occurrence of $k$ in $\min\{\log_2(N), kC'(\theta_{\rm GCN})\}$.}. 
\end{theoreminformal}

\paragraph{Metric embedding tools for transductive learning guarantees}

We introduce a new approach for establishing transductive learning guarantees on random graphs.
The central idea is to represent a given large, perhaps high-dimensional, input graph in $(\mathbb{R}^m, d_{\infty})$, where $d_{\infty}$ denotes the metric induced by the $\ell^{\infty}$-norm, and $m=1,2$.
Specifically, these representations approximate the geometry of the original graph through fractal embeddings, known as non-Lipschitz bi-H\"older maps, which preserve (selected) fractional powers of graph distances within controlled distortion.
In constructing these embeddings, we draw on recent advances in metric embedding~\citep{OferLowDimEmbeddingDoublingSpaces} and classical results from metric geometry~\citep{schoenberg1938metric,assouad1983plongements}.
Once the graph is embedded in low dimensions, we reformulate the transductive learning problem as a concentration of empirical measure statement in the $1$-Wasserstein distance.
A key observation is that an empirical measure, from a Borel measure, concentrates at the nonparametric rate $\mathcal{O}(1/N^{1/2})$ in dimension one, with only a logarithmic slowdown, $\mathcal{O}(\log(N)/N^{1/2})$ in dimension two. 
The optimal representation dimension $m=1,2$ is chosen adaptively to minimize the generalization gap as a function of $N$, taking into account $\min\{\log_2(N),C'(k,\theta_{\rm GCN})\}$ as suggested by \eqref{eq:rates}, and this choice is determined in the concluding stage of our analysis.

\subsection{Related works and frameworks}
\label{s:Intro__ss:RelatedWorks}

\paragraph{Multiple learning regimes} 
Standard probably approximately correct (PAC) learning theory aims to control \textit{generalization}, defined as the difference between performance on training (in-sample) data and unseen test (out-of-sample) data. Classical bounds are established for a single, fixed learning regime and take the following form:
\begin{equation}
\label{eq:PAC_Learning__structure}
        \frac{C}{N^{1/2}} + \frac{(\log(1/\delta))^{1/2}}{N^{1/2}}, 
\end{equation}
where $N$ is the sample size (e.g., the number of sampled graph nodes), $\delta\in (0,1)$ is the failure probability, and $C>0$ depends on the cardinality or metric entropy of the hypothesis class~\citep{arora2018stronger,bartlett2021failures,hou2023instance}. 
The focus is typically to determine the sharpest possible constant $C$; see, e.g., \citep{kontorovich2019exact}.
Such bounds, however, have an inherently single-phase character: their asymptotic behaviour as $N\to\infty$ dominates, often yielding vacuous guarantees at practical sample sizes; see, e.g., \citep{dziugaite2017computing}.
Our analysis instead establishes a two-phase learning regime that adapts simultaneously to the sample size $N$ and the number of verified nodes $k$, scaled by the graph learner structure $\theta_{\rm GCN}$. 
The resulting flexibility yields non-vacuous bounds even for moderate $N$, while still achieving the asymptotically optimal PAC rate $\mathcal{O}(N^{-1/2})$.
Moreover, our bound sharpens to $C\log_2(N)N^{-1/2}$ prior to the phase transition, reminiscent of the sample-size enlargement effect in information theory \citep{JIANminimax_TIT_2015,lepski1999estimation,cai2011testing}
and double descent in modern statistical learning \citep{belkin2020two,bach2024high,shi2024homophily}. 
In a broader context, the multi-phase behavior parallels phenomena in AI statistics, such as phase transitions in differential privacy \citep{zamanlooy2023strong}, spectral separation in learning \citep{yarotsky2025corner}, and expressivity gaps in neural networks \citep{mhaskar2017and,NEURIPS2020_979a3f14}.

\paragraph{Transductive learning guarantees under common noise}  
In our second main result, Theorem~\ref{thrm:Main_Result__random}, the \textit{true} and \textit{empirical risks} (see \eqref{eqdef:truerisk} and \eqref{eqdef:emprisk}, respectively) are evaluated conditionally on a single draw of a random $k\times d_{\rm in}$ feature matrix $\mathbf{X}$ and a random graph $\mathbf{G}$. Consequently, the risks are random and share a \textit{common source of randomness}. This mirrors \textit{mean-field behaviour with common noise}~\citep{CarmonaDelarueLackerMFGCN_2016,Dylan2022}, where correlated randomness complicates asymptotic analysis.  
Further, the input randomness induces additional probabilistic challenges that do not arise in classical PAC learning and thus prevent the direct use of standard tools from empirical process theory (e.g.~\citep[Pages 16-28]{van1996weak}) and uniform central limit theorems (e.g.~\citep[Chapter~6]{shalev2014understanding}).

\paragraph{Tools from metric embedding theory}
As mentioned, we cast transductive learning as a measure concentration problem in a Euclidean space $\mathbb{R}^m$, following the framework recently developed in~\citep{kratsios2024tighter}.
This introduces an inherent trade-off: higher-dimensional representations yield learning bounds with smaller constants but slower convergence, whereas lower-dimensional ones give faster convergence at the expense of larger constants.
Exploiting this trade-off allows us to identify multiple learning regimes by adaptively choosing the embedding dimension $m$ as a function of the sample size and other geometric invariants of the underlying graph.
Our approach departs from~\citep{kratsios2024tighter} in subtle yet critical ways. Specifically, we embed a \textit{snowflaked} version of the underlying graph into $\mathbb{R}^m$ equipped with the $\ell^{\infty}$ norm. For a snowflake degree $\alpha \in (0,1)$, the $\alpha$-snowflaked metric raises each original distance to the power $\alpha$\footnote{The extremal cases are $\alpha=0$, yielding the uniform metric where all nonzero distances equal one, and $\alpha=1$, which recovers the original metric.}.  
Focusing on the $\ell^{\infty}$ norm leverages the Kuratowski embedding theorem \citep[page 99]{heinonen2001lectures}, which ensures $(\mathbb{R}^m, \ell^{\infty})$ contains isometric copies of every $k$-point metric space for $1 \le k \le m$.
Moreover, for finite doubling metric spaces, including our (random) graphs, snowflaking combined with $\ell^{\infty}$-distance allows embeddings whose target dimension and distortion\footnote{The distortion of a bi-Lipschitz embedding quantifies how much it stretches versus contracts distances.} depend solely on the snowflake degree and doubling constant, but not on the cardinality of the space (see \citep[Theorem 3]{OferLowDimEmbeddingDoublingSpaces}); such independence is key in our analysis.

\subsection{Organization}

Section~\ref{s:Prelims} reviews the necessary background and consolidates the notation and terminology required to state our main results. Section~\ref{s:Main} presents these results, distinguishing between the deterministic setting with a fixed graph and node-level features (Theorem~\ref{thrm:main_result__deterministic}) and the common noise setting, where both the training and testing sets share a single realization of a random graph and feature matrix (Theorem~\ref{thrm:Main_Result__random}).
We then illustrate the applicability of our results to transductive learning with graph convolutional networks, both for deterministic graphs with no isolated vertices (Corollary~\ref{cor:Transductive_GCNNs}) and for graphs drawn once from an Erd\H{o}s-R\'enyi random graph (Corollary~\ref{cor:Transductive_GCNNs__random}).
Section~\ref{s:toolsandproofs} outlines our key proof techniques and introduces technical tools of potential independent interest. In particular, this includes a concentration of measure result in the $1$-Wasserstein distance for finite metric spaces via metric snowflaking (Proposition~\ref{prop:New_Convergence__SuperAssouad}).
Following the exposition in Section~\ref{s:toolsandproofs}, detailed proofs of Theorem~\ref{thrm:main_result__deterministic} and Theorem~\ref{thrm:Main_Result__random} are provided in Sections~\ref{sec:thm1proof} and~\ref{sec:thm2proof}, respectively.
Proofs of technical tools introduced in Section~\ref{s:toolsandproofs} as well as further necessary backgrounds are given in the Appendix. 
Lastly, Appendix~\ref{s:AuxiliaryResults__ss:Proofs} provides various upper-bound estimates of the metric doubling constant for graphs with diameter at most $2$. These results are of independent interest and may be useful to researchers working on metric embedding theory.

\section{Preliminaries}
\label{s:Prelims}
We review the essential preliminary concepts and conventions.  
We write $\mathbb{N}$ to denote the set of natural numbers and $\mathbb{R}_{\geq 0}$ to denote the set of nonnegative reals.
For a finite set $V$, we denote by $\#V$ its cardinality.
For a linear operator $W: \R^m\to\R^n$, we define $\|W\|_{\rm op} \eqdef \sup_{x\in \R^m}\|W x\|_2/\|x\|_2$ to be its operator norm, where $\|\cdot\|_2$ denotes the Euclidean norm. 
Lastly, we use the analyst's constant notations $C,c$, which are allowed to change value from one instance to the next.

\paragraph{Graphs}  Let $G=(V,E)$ denote a graph with vertex set $V$ and edge set $E \subset V \times V$. 
We restrict to the case of $G$ being a finite, simple (undirected) graph with no isolated vertices. Suppose $\# V=k$, for $k \in \mathbb{N}$. We associate with $G$ a graph adjacency matrix $A_G \in \mathbb{R}^{k \times k}$, where $[A_{G}]_{i,j}=1$ if and only if $\{v_i,v_j\}\in E$, and otherwise $[A_{G}]_{i,j}=0$. The degree of $v_i \in V$ is given to be ${\rm deg}(v_i) \eqdef \sum_{j=1}^k [A_G]_{ij}$, while the maximal and minimal graph degrees are defined respectively by ${\rm deg}_{+}(G)\eqdef \max_{v\in V} {\rm deg}(v)$ and ${\rm deg}_{-}(G)\eqdef \min_{v\in V} {\rm deg}(v)$. 
A graph $G$ has no isolated vertices if ${\rm deg}_{-}(G)\geq 1$.
We denote by $D_G$ the degree matrix of $G$, i.e. $[D_G]_{ij} \eqdef \mathbbm{1}_{\{i= j\}} {\rm deg}(i)$. 

\paragraph{Metric spaces}
Let $(\mathscr{X},d_{\mathscr{X}})$ denote a metric space; whenever clear from the context, we simply write $\mathscr{X}$ in place of $(\mathscr{X},d_{\mathscr{X}})$. The diameter of $(\mathscr{X},d_{\mathscr{X}})$ is defined to be ${\rm diam}(\mathscr{X}) \eqdef \sup_{x,x'\in\mathscr{X}} d_{\mathscr{X}}(x,x')$.
A (closed) ball of radius $r\geq 0$, centred at $x\in\mathscr{X}$ is given as, 
\begin{equation*}
    B(x,r) \eqdef \{y\in\mathscr{X}: d_{\mathscr{X}}(x,y)\leq r\}.
\end{equation*}
The $k$-fold Cartesian product $(\mathscr{X}^{k}, d_{\mathscr{X}^{k}})$ of $(\mathscr{X},d_{\mathscr{X}})$ is a metric space equipped with the product metric
\begin{equation} \label{eqdef:inftyprodmetric}
    d_{\mathscr{X}^{k}}((x_v)_{v=1}^k,(x'_{v})_{v=1}^k)
    \eqdef 
    \max_{v=1,\dots,k} d_{\mathscr{X}}(x_v,x'_{v}).
\end{equation}
Under this metric, ${\rm diam}(\mathscr{X}^{k}) = {\rm diam}(\mathscr{X})$.
Let $(\mathscr{Y},d_{\mathscr{Y}})$ be another metric space.
Then similarly, the Cartesian product $\mathscr{X}\times\mathscr{Y}$ is a metric space with the metric
\begin{equation*} 
    d_{\mathscr{X}\times\mathscr{Y}} ((x,y),(x',y'))
    \eqdef 
    \max\{ d_{\mathscr{X}}(x,x'), d_{\mathscr{Y}}(y,y') \}.
\end{equation*}
We say that $(\mathscr{X},d_{\mathscr{X}})$ is doubling with the doubling constant $\mathtt{M}\in\mathbb{N}$, if for every $r\geq 0$ and every $x\in\mathscr{X}$, the closed ball $B(x,r)$ can be covered by some $\mathtt{M}$ closed balls $B(x_1,r/2),\dots,B(x_{\mathtt{M}},r/2)$, i.e.
\begin{equation*}\label{eq::doublingconstantM}
    B(x,r) \subset \bigcup_{i=1}^{\mathtt{M}} B(x_i,r/2),
\end{equation*}
and if $\mathtt{M}$ is the smallest such number. 
Below, we present three examples of prototype metric spaces that will be our main focus.

\begin{example}
\label{exam:Euclideanmetric}
When $\mathscr{X}$ is a subset of a Euclidean space $\mathbb{R}^d$, we endow it with the metric $d_{\infty}$ induced by the $\ell^{\infty}$-norm; namely, $d_{\infty}(x,x') \eqdef \|x-x'\|_{\infty}$,
where in dimension one, $\|\cdot\|_{\infty}$ is simply the absolute value $|\cdot|$.
It is readily verified that $(\mathscr{X},d_{\infty})$ is doubling with the doubling constant $2^d$. 
\end{example}

\begin{example}\label{exam:graphmetricspace}
Let $G=(V, E)$ be a finite, simple graph with vertex set $V$ and edge set $E$, equipped with the \textit{shortest path length} metric $d_G$, forming the \textit{graph metric space} $(G,d_G)$.
Thus, for example, when $G$ is disconnected, ${\rm diam}(G)=\infty$.
Provided that $G$ is non-singleton, it can be checked that $(G,d_G)$ is doubling with the doubling constant $2\leq \mathtt{M} \leq\#V<\infty$. 
\end{example}

\begin{example}\label{exam:coordinates}
Let $[k]\eqdef\{1,\dots,k\}$ be an index set, which can be viewed as representing the vertex set of a finite, simple graph $G$.
We equip $[k]$ with the metric $d_{[k]}=d_G$; note that such a metric choice is inherently determined by the prior selection of $G$. Consequently, $([k], d_G)$ is isometric with $(G,d_G)$, and the distinction between them is only formal.
\end{example}

\paragraph{Graph learners}
We introduce a broad class of hypotheses that process graph and node features, to which our analysis applies.
Let $\mathcal{G}_k$ denote the collection of simple (undirected) graphs on the vertex set $[k] =\{1,\dots,k\}$. 
Let $ E_{\rm in}^{k}$ and $ E_{\rm out}^{k}$ denote a \textit{feature space} and a \textit{label space} defined on the vertex set, respectively, where $ E_{\rm in}\subset\mathbb{R}^{d_{\rm in}}$, $d_{\rm in}\in\N$, and $ E_{\rm out}\subset\mathbb{R}$. 
For $x\in E_{\rm in}^{k}$ (resp. $y\in E_{\rm out}^{k}$) and $v\in [k]$, let $\pi_v(x)$ (resp. $\pi_v(y)$) denote the projection of $x$ (resp. $y$) onto its $v$-th coordinate. 
Fix $G\in\mathcal{G}_k$. We equip $[k]$ with the shortest graph length metric $d_G$. For $\mathtt{B}>0$, we denote by $\mathcal{F}_{\mathtt{B}}$ the class of hypotheses $f: E_{\rm in}^k\to E_{\rm out}^k$ that are $\mathtt{B}$-Lipschitz with respect to both the input features and the metric space $([k],d_G)$. 
That is, if $f\in \mathcal{F}_{\mathtt{B}}$, then 
for $x, x'\in E_{\rm in}^k$,
\begin{equation} \label{eq:Liprequire1}
    \|f(x) - f(x')\|_{\infty} \leq \mathtt{B}\|x-x'\|_{\infty},
\end{equation}
and for every $x\in E_{\rm in}^k$ and $v,v'\in [k]$\footnote{Since $E_{\rm out}$ is bounded and $d_G(v,v') \geq 1$, \eqref{eq:Liprequire2a} is equivalent to, $|\pi_v(f(x)) - \pi_{v'}(f(x))| \leq \mathtt{B}'$, for some $\mathtt{B}'\geq \mathtt{B}$.}, 
\begin{equation} \label{eq:Liprequire2a}
    |\pi_v(f(x)) - \pi_{v'}(f(x))| \leq \mathtt{B} d_G(v,v').
\end{equation}
A class of graph learners satisfying \eqref{eq:Liprequire1} and \eqref{eq:Liprequire2a} can be obtained from (generalized) GCNs, which are the focus of our study, defined below.

\paragraph{GCNs}
Let $G\in \mathcal{G}_k$. Let $A_G \in \mathbb{R}^{k\times k}$ be its adjacency matrix and $D_G$ be its degree matrix. Let
\begin{equation*}
    \Delta_G
    \eqdef 
    I_k - D_G^{-1/2} A_G D_G^{-1/2},
\end{equation*} 
be its (normalized) graph Laplacian.
For $t\in\mathbb{N}$, let $(\Delta_G)^t$ be the $t$-power of $\Delta_G$. We consider the following GCN model; see~\citep[Chapter~5.3]{ma2021deep}. 

\begin{definition} 
\label{defn:GGCN}
Let $k\in\N$, and let $\mathcal{G}_k$ be the set of simple graphs on $[k]$.
Let $L,t, d_{\rm in}\in\N$ and $d_{\rm out}=1$. 
Let $\beta_1,\dots,\beta_L>0$.
For $l=0,1,\dots,L$, let $d_l\in\N$, with $d_0\eqdef d_{\rm in}$, $d_L\eqdef d_{\rm out}=1$.
Let $E_{\rm in}\subset\R^{d_{\rm in}}$ and $E_{\rm out}\subset\R$.
For $l=1,\dots,L-1$, let $W_l\in\R^{d_l\times d_{l+1}}$ be given weight matrices, with $\|W_l\|_{\rm op}\le \beta_l$.
Let $\sigma:\mathbb{R}\to \mathbb{R}$ be a given $1$-Lipschitz activation function. 
The class $\mathcal{F}_{\rm GCN}$ on $\mathcal{G}_k$ consists of maps 
\begin{equation*}
    f: \mathcal{G}_k\times E_{\rm in}^{k} \to E_{\rm out}^{k}\subset \mathbb{R}^{d_{\rm out}\times k}
\end{equation*}
that are defined by generalized GCNs whose architecture is specified by $t$-hop convolution, activation $\sigma$, and \textit{network parameters} $(W_1,\dots,W_L)$, and whose \textit{network size} is given by $(\beta_1,\dots,\beta_L)$. These maps admit the following iterative representation. For each $G\in\mathcal{G}_k$ and $x\in E_{\rm{in}}^k$, let $f(G,x)\eqdef H_L \eqdef W_L H_{L-1}$, where 
\begin{align} \label{eq:GCNcompute}
    H_{l+1} \eqdef \mathfrak{L}_{l+1}(H_l) \quad\text{ for }\quad l=0,1,\dots,L-2, \quad\text{ and }\quad H_0 \eqdef x.
\end{align}
Here in \eqref{eq:GCNcompute}, $\mathfrak{L}_l(\tilde{x}) \eqdef \sigma\bullet (W_l((\Delta_G)^t \tilde{x}^{\top})^{\top})$, for $\tilde{x}\in \mathbb{R}^{d_l\times k}$, where $\bullet$ denotes a component-wise application. 
\end{definition}

Corollary~\ref{cor:Transductive_GCNNs} shows that, when restricted to learning on a graph $G$ with no isolated vertices, this class of generalized GCNs belongs to the function class $\mathcal{F}_{\mathtt{B}}$; see \eqref{eq:GCNLip}.

\section{Setup and main results}
\label{s:Main}

\subsection{Transductive learning setup} \label{sec:TL}

Let $\mathcal{G}_k$ be the collection of simple graphs on $[k]$. We first consider the case where $G\in\mathcal{G}_k$ is deterministic. We assume $G$ has no isolated vertices.
Let $([k], d_G)$ be the associated metric space, with $d_G$ denoting the shortest path length metric of $G$.
Let $ E_{\rm in}^{k}$ and $ E_{\rm out}^k$ be respectively the feature and label spaces on $[k]$, where $ E_{\rm in}\subset\mathbb{R}^{d_{\rm in}}$ and $ E_{\rm out}\subset\mathbb{R}$ are bounded.
Let $\mathcal{F}_{\mathtt{B}}$ denote the class of functions $f: E_{\rm in}^k\to E_{\rm out}^k$ satisfying \eqref{eq:Liprequire1}, \eqref{eq:Liprequire2a}.
Let $f^{\star}\in \mathcal{F}_{\mathtt{B}}$ be a target function. We consider the following TL problem induced by $f^{\star}$ and a fixed $x\in E_{\rm in}^k$.
Let $\mathbb{P}_{[k]}$ be a probability measure on $[k]$, and let\footnote{This means that $\mathbb{P}$ is the push-forward of $\mathbb{P}_{[k]}$ under $(\mathbbm{1}\times g_{[k]})$.} 
$\mathbb{P} \eqdef (\mathbbm{1}\times g_{[k]})_{\#}\mathbb{P}_{[k]}$, where $g_{[k]}(v)\eqdef \pi_v(f^{\star}(x))$.
Let us be supplied with independent random samples
\begin{equation*} 
    (V_1,Y_1),\dots,(V_N,Y_N)\sim\mathbb{P},
\end{equation*}
taking values on $[k]\times E_{\rm out}$; that is, $Y_i = g_{[k]}(V_i)=\pi_{V_i}(f^{\star}(x))$. 
Let $\ell: E_{\rm out}\times E_{\rm out} \rightarrow \mathbb{R}_{\geq 0}$ be a $\mathtt{B}_{\ell}$-Lipschitz loss function, 
\begin{equation} \label{ellLip}
    |\ell(y,z) - \ell(y',z')| \leq \mathtt{B}_{\ell} \max\{|y-y'|, |z-z'|\},
\end{equation}
and let $\ell_{1/2}: E_{\rm out}\times E_{\rm out} \rightarrow \mathbb{R}_{\geq 0}$ be its $1/2$-snowflaked version; that is, $\ell_{1/2}(y,z) \eqdef \ell(y,z)^{1/2}$. 
Using this snowflaked loss, we take the empirical risk to be
\begin{equation} \label{eqdef:emprisk}
    \mathcal{R}_{G,x}^N(f)  \eqdef \frac1{N} \sum_{n=1}^N
    \ell_{1/2}(\pi_{V_n}(f(x)),Y_n),
\end{equation}
and the corresponding true risk to be
\begin{equation} \label{eqdef:truerisk}
    \mathcal{R}_{G,x}
    (f) \eqdef \mathbb{E}_{(V,Y)\sim\mathbb{P}} \big[\ell_{1/2}(\pi_V (f({x})),Y)\big].
\end{equation}
The worst-case discrepancy between these two quantities is captured by the \textit{transductive generalization gap} over the hypothesis class $\mathcal{F}_{\mathtt{B}}$
\begin{equation*}
    \sup_{f\in\mathcal{F}_{\mathtt{B}}} 
    \big|\mathcal{R}_{G,x}(f) - \mathcal{R}_{G,x}^N(f) \big|.
\end{equation*}
Estimating this gap broadly defines our TL problem. We will consider two settings: the deterministic case, where both the graph and given feature are deterministic, and the common noise case, where both the graph and feature are random.

\subsection{A transductive learning result on deterministic graphs}
\label{s:Main__ss:Deterministic}

Our first main result addresses the transductive learning problem in the deterministic case. We adopt the setting introduced in Section~\ref{sec:TL}. 

\begin{theorem} 
\label{thrm:main_result__deterministic}
Let $k,N\in \mathbb{N}$ such that $k\geq 2$, $N\ge 4$. 
Let
\begin{equation} \label{eq:rates}
    \mathtt{r}_1(N) \eqdef \frac{\log_2(N)}{N^{1/2}} \quad\text{ and }\quad
    \mathtt{r}_2(N) \eqdef 
    \frac{    
    k
    ({\rm diam}(G) + {\rm diam}( E_{\rm out}))^{1/2}
    }{N^{1/2}},
\end{equation}
and for each $\delta\in (0,1)$, let
\begin{equation} \label{eq:tail}
    \mathtt{t}(N,\delta) \eqdef \frac{(3\log_2(2/\delta)({\rm diam}(G) + {\rm diam}( E_{\rm out})))^{1/2}}{N^{1/2}}.
\end{equation}
Then it holds with probability at least $1-\delta$ that
\begin{multline*} 
    \sup_{f\in\mathcal{F}_{\mathtt{B}}} |\mathcal{R}_{G,x}(f) - \mathcal{R}^N_{G,x}(f)|
    \leq (2\mathtt{B}_{\ell} \mathtt{B})^{1/2}\Big(({\rm diam}(G) + {\rm diam}( E_{\rm out}))^{1/2}
    \min\{
    4\mathtt{r}_1(N), 48\mathtt{r}_2(N)
    \} + \mathtt{t}(N,\delta)\Big).
\end{multline*}
\end{theorem}

\paragraph{Application: transductive learning guarantees for GCNs.}

We apply Theorem~\ref{thrm:main_result__deterministic} to the case where the hypothesis class consists of common GCN models given in Definition~\ref{defn:GGCN}.

\begin{corollary} 
\label{cor:Transductive_GCNNs}  
Let $k, N\in \mathbb{N}$ such that $k\geq 2$, $N\geq 4$. 
Let $G\in\mathcal{G}_k$ such that ${\rm deg}_{-}(G)\geq 1$. 
Let the hypothesis subclass $\mathcal{F}_{\rm GCN}$ be given in Definition~\ref{defn:GGCN} with 
$E_{\rm in}$, $E_{\rm out}$ bounded. 
Then for each $\delta\in (0,1)$, it holds with probability at least $1-\delta$ that
\begin{multline*} 
    \sup_{f\in\mathcal{F}_{\rm GCN}} |\mathcal{R}_{G,x}(f) - \mathcal{R}^N_{G,x}(f)|
    \leq (2\mathtt{B}_{\ell} \mathtt{B})^{1/2}\Big(({\rm diam}(G) + {\rm diam}( E_{\rm out}))^{1/2}
    \min\{
    4\mathtt{r}_1(N), 48\mathtt{r}_2(N)
    \} + \mathtt{t}(N,\delta)\Big).
\end{multline*}
where 
\begin{equation} \label{eq:GCNLip}
    \mathtt{B} \eqdef \max\Big\{d_{\rm in}^{1/2} \bigg(1 + \frac{(k-1)^{1/2}}{{\rm deg}_{-}(G)^{1/2}}\bigg)^{tL} 
    \prod_{l=1}^{L} 
    \beta_l
    ,
    \, {\rm diam}(E_{\rm out})\Big\}.
\end{equation}
\end{corollary}
We emphasize that each $f \in \mathcal{F}_{\rm GCN}$ takes as input both a graph and node features. However, once $G$ is fixed, $f(G,\cdot)$ depends only on the features, and the TL problem in Corollary~\ref{cor:Transductive_GCNNs} is interpreted in this setting. For notational convenience, we continue to write $f \in \mathcal{F}_{\rm GCN}$. 
Importantly, Corollary~\ref{cor:Transductive_GCNNs} provides the Lipschitz regularity $\mathtt{B}$ of $f\in\mathcal{F}_{\rm GCN}$, as specified in \eqref{eq:GCNLip}, which plays a central role in Theorem~\ref{thrm:Main_Result__random} below.\\

\noindent\textit{Proof.} See Appendix~\ref{appx:Transductive_GCNNs}.

\subsection{A transductive learning result under shared input randomness}
\label{s:Main__ss__Random}

Throughout this section, boldface and capitalization are used exclusively for the graph $\mathbf{G}$ and input feature $\mathbf{X}$ when these objects carry randomness, distinguishing this noisy setting from the previous deterministic one.
Other sources of randomness, such as sampling, are not affected by this convention.
To formalize the TL result in this setting, we introduce the probabilistic setup and necessary assumptions, beginning with our graph models. 

\begin{assumption}[Admissible random graph models]
\label{ass:Super-Density}
For every $k \in \mathbb{N}$, let $\mathcal{U}_k \subset \mathcal{G}_k$ be a nonempty set of simple graphs on the vertex set $[k]$.  
\begin{enumerate}
\item[(i)] 
    We say that the collection $\{\mathcal{U}_k\}_{k\in\N}$ is \textit{admissible} if there exists a sequence $\{c_k\}_{k\in\N}$ of positive numbers such that, for each $G_k\in\mathcal{U}_k$, we have ${\rm diam}(G_k)\leq 2$ and ${\rm deg}_{-}(G_k)\geq c_k$.\footnote{Since a graph with diameter at most $2$ is necessarily connected, we could indeed choose $c_k = 1$ for all $k\in \mathbb{N}$. However, depending on the family $\{\mathcal{U}_k\}_{k\in\N}$, this choice might not be optimal.}
\item[(ii)]  
    We say that a collection of random graphs $\{\mathbf{G}_k\}_{k\in\N}$ is \textit{admissible} with respect to an \textit{admissible} $\{\mathcal{U}_k\}_{k\in\N}$ if $\lim_{k\rightarrow\infty}\mathbb{P}(\mathbf{G}_k \in \mathcal{U}_k)=1$.
\end{enumerate}
The condition (ii) implies that for $k\in\N$ sufficiently large, the event $\mathbf{G}_k\in\mathcal{U}_k$ happens with high probability.
When $k$ is clear from the context, we write $\mathbf{G} = \mathbf{G}_k$ and $G = G_k$.
\end{assumption}

In particular, for the Erd\H{o}s-R\'enyi random graph $\mathbf{G}=\mathbf{G}(k,p(k))$ with $p(k) = (c\log (k)/k)^{1/2} \in (0,1)$, we may take, $c_k=(ck\log (k))^{1/2}$. That is, ${\rm deg}_-(\mathbf{G})\geq (ck\log (k))^{1/2}$ with high probability; see Lemma~\ref{lem:Degree}.

Next, we recall the hypothesis class $\mathcal{F}_{\rm GCN}$ of GCNs given in Definition \ref{defn:GGCN}, and considered in Corollary~\ref{cor:Transductive_GCNNs}.
Here, as the input feature $\mathbf{X}$ is allowed to be noisy, we impose that its entries are bounded almost surely.

\begin{assumption}[Admissible features] 
\label{ass:Features}
We observe a single \textit{random feature matrix} from the family $\{\mathbf{X}_k\}_{k\in\mathbb{N}}$, where each $\mathbf{X}_k$ is a random $d_{\rm in}\times k$ matrix whose columns lie in $[-M,M]^{d_{\rm in}}$ with probability one, for some absolute $M\geq 1/2$ (effectively, $E_{\rm in}=[-M,M]^{d_{\rm in}}$). When $k$ is clear from context, we write $\mathbf{X} = \mathbf{X}_k$. 
\end{assumption}

We present our second main result, addressing the TL problem from Section~\ref{sec:TL} in the presence of shared randomness from single observations of both the random feature matrix and graph, for the hypothesis class $\mathcal{F}_{\rm GCN}$.

\begin{theorem}
\label{thrm:Main_Result__random}
Let $k,N\in \mathbb{N}$ such that $k\geq 2$, $N\ge 4$.
Let the hypothesis class $\mathcal{F}_{\rm GCN}$ be given in Definition~\ref{defn:GGCN}, with a random input graph $\mathbf{G}$ satisfying Assumption~\ref{ass:Super-Density}(ii) and an input feature $\mathbf{X}$ satisfying Assumption~\ref{ass:Features}. 
Let
\begin{equation*} 
    \mathtt{r}_1(N) \eqdef \frac{\log_2(N)}{N^{1/2}} \quad\text{ and }\quad
    \mathtt{r}_2(N) \eqdef 
    \frac{    
    k
    (2 + \mathtt{D})^{1/2}
    }{N^{1/2}},
\end{equation*}
and for each $\delta\in (0,1/2)$, let
\begin{equation*} 
    \mathtt{t}(N,\delta) \eqdef \frac{(3\log_2(2/\delta)(2 + \mathtt{D}))^{1/2}}{N^{1/2}},
\end{equation*}
where
\begin{equation} \label{eq:matD}
    \mathtt{D} \eqdef 2Md_{\rm in}^{1/2} \big(1 + c_k^{-1/2}(k-1)^{1/2}\big)^{tL} 
    \prod_{l=1}^{L} 
    \beta_l.
\end{equation}
Then, for sufficiently large $k\in\N$, depending on $\delta$, the following holds with probability at least $1-2\delta$
\begin{align} \label{eq:thmrate}
    \sup_{f\in\mathcal{F}_{\rm GCN}} |\mathcal{R}_{\mathbf{G},\mathbf{X}}(f) - \mathcal{R}^N_{\mathbf{G},\mathbf{X}}(f)|
     \leq (2\mathtt{B}_{\ell} \mathtt{D})^{1/2}
     \big(
         (2+\mathtt{D})^{1/2} \min\{
         4\mathtt{r}_1(N), 48\mathtt{r}_2(N)\} + \mathtt{t}(N,\delta)
     \big).
\end{align}
\end{theorem}

\noindent
\textit{Proof.} See Section~\ref{sec:thm2proof}.

\begin{remark}
One can relax Assumption~\ref{ass:Features} by assuming that the columns of each $\mathbf{X}_k$ are independent, sub-Gaussian random vectors, with mean zero and sharing the same positive definite covariance matrix. This effectively requires that the columns of the feature matrices be standardized. Then due to the strong concentration properties of sub-Gaussian vectors, Theorem~\ref{thrm:Main_Result__random} would still hold, up to an additional concentration probability term.
\end{remark}

\paragraph{Application: transductive learning guarantees for GCNs with common noise induced by an Erd\H{o}s-R\'{e}nyi graph.}
The Erd\H{o}s-R\'{e}nyi model $\mathbf{G}=\mathbf{G}(k,p)$ is a random graph on $k$ nodes where each of the $\binom{k}{2}$ possible edges appears independently with probability $p=p(k)$.
We apply Theorem~\ref{thrm:Main_Result__random} with the input graph given by $\mathbf{G}=\mathbf{G}(k,p(k))$, where for $C>2$, and sufficiently large $k$, 
we let $p(k)=(C\log (k)/k)^{1/2}\in (0,1)$.

\begin{corollary}
\label{cor:Transductive_GCNNs__random}
Let $k,N\in \mathbb{N}$ such that $k\geq 2$, $N\ge 4$.
Let the hypothesis class $\mathcal{F}_{\rm GCN}$ be given in Definition~\ref{defn:GGCN}, with a random input Erd\H{o}s-R\'{e}nyi random graph $\mathbf{G}=\mathbf{G}(k,p(k))$, where $p(k)=(C\log (k)/k)^{1/2}$, and an input feature $\mathbf{X}$ satisfying Assumption~\ref{ass:Features}.
Let $\delta\in (0,1/2)$.
Then, for sufficiently large $k\in\N$, depending on $\delta$, the following holds with probability at least $1-2\delta$
\begin{align*}
    \sup_{f\in\mathcal{F}_{\rm GCN}} |\mathcal{R}_{\mathbf{G},\mathbf{X}}(f) - \mathcal{R}^N_{\mathbf{G},\mathbf{X}}(f)|
     \leq (2\mathtt{B}_{\ell} \mathtt{D})^{1/2}
     \big(
         (2+\mathtt{D})^{1/2} \min\{
         4\mathtt{r}_1(N), 48\mathtt{r}_2(N)\} + \mathtt{t}(N,\delta)
     \big).
\end{align*}
Here, 
\begin{equation} \label{eq:updatedD}
    \mathtt{D} \eqdef 2Md_{\rm in}^{1/2} \Big(1 + \Big(\frac{c(k-1)}{k\log (k)}\Big)^{1/2}\Big)^{tL} 
    \prod_{l=1}^{L} 
    \beta_l,
\end{equation}
and $c$ is an absolute constant.
\end{corollary}

\noindent
\textit{Proof.} See Appendix \ref{app:proof_of_corollary2}.

\section{Main technical tools}
\label{s:toolsandproofs}

\subsection{Main technical tool for Theorem~\ref{thrm:main_result__deterministic}}

Theorem~\ref{thrm:main_result__deterministic} builds on a concentration inequality for empirical measures on doubling metric spaces, adapted from \citep{fournier2015rate,Kloeckner_2020Benoit} and expressed in terms of the (H\"older) Wasserstein distance. 
Stated as Proposition~\ref{prop:New_Convergence__SuperAssouad} below, this result enables the application of Assouad's metric embedding theory~\citep{assouad1983plongements,david2013non,naor2012assouad} to doubling metric spaces.

Let $\alpha\in (0,1]$. 
The $\alpha$-H\"older Wasserstein distance between two probability measures $\mu$, $\nu$ on $\mathscr{X}$ 
is given by (see \citep[Definition 9]{hou2023instance})\footnote{Definition \eqref{eqdef:Wassdistance} is inspired by the fact that setting $\alpha=1$ recovers the dual definition \citep[Remark 6.5]{villani2009optimal} of the Wasserstein $\mathcal{W}_1$ transport distance \citep[Definition~6.1]{villani2009optimal}.} 
\begin{equation} \label{eqdef:Wassdistance}
    \mathcal{W}_{\alpha}(\mu,\nu)
    \eqdef 
    \sup_{f\in {\rm H}(\alpha,\mathscr{X},1)}\, 
    \mathbb{E}_{X\sim \mu}[f(X)] - \mathbb{E}_{Y\sim \nu}[f(Y)],
\end{equation}
where, for $\mathtt{B}\geq 0$, ${\rm H}(\alpha,\mathscr{X},\mathtt{B})$ denotes the set of real-valued $\alpha$-H\"older continuous functions $f$ on $\mathscr{X}$ satisfying 
\begin{equation} \label{stapleton}
    |f(x)-f(x')|\leq \mathtt{B}d_{\mathscr{X}}(x,x')^{\alpha},
\end{equation}
for every $x,x'\in\mathscr{X}$.
Note, when $\alpha=1$, ${\rm H}(1,\mathscr{X},\mathtt{B})={\rm Lip}(\mathscr{X},\mathtt{B})$, the set of real-valued $\mathtt{B}$-Lipschitz continuous functions on $\mathscr{X}$. Note further that the definition \eqref{stapleton}, and thus \eqref{eqdef:Wassdistance}, depends on the metric choice. For example, if $\mathscr{X}$ have been equipped with the \textit{snowflaked metric} $d_{\mathscr{X}}^{\alpha}$ -- that is, every distance is raised to the power $\alpha$ -- then \eqref{stapleton} would describe a $\mathtt{B}$-Lipschitz function on $(\mathscr{X},d_{\mathscr{X}}^{\alpha})$. 
Throughout, we take care to specify the metrics in use.

\begin{proposition}
\label{prop:New_Convergence__SuperAssouad}
Let $(\mathscr{X},d_{\mathscr{X}})$ be a $k$-point doubling metric space, with $k\geq 2$ and the doubling constant 
$\mathtt{M}\geq 2$. 
Assume $d_{\mathscr{X}}(x,x')\geq 1$ for all $x\not=x'\in\mathscr{X}$.
Let $\mu$ be a probability measure on $\mathscr{X}$, and let $\mu^N$ be its associated empirical measure.
Let
\begin{equation*} 
    \mathtt{r}_1(N) \eqdef \frac{\log_2(N)}{N^{1/2}}, \quad
    \mathtt{r}_2(N) \eqdef 
    \frac{    
    k
    {\rm diam}(\mathscr{X})^{1/2}
    }{N^{1/2}}, \quad
    \mathtt{r}_3(N) \eqdef \frac{1}{N^{1/\lceil 4\mathtt{M}^{5+\log_2(5)} \rceil}}, 
\end{equation*}
and for each $\delta\in (0,1)$, let
\begin{equation*} 
    \mathtt{t}(N,\delta) \eqdef \frac{(3\log_2(2/\delta){\rm diam}(\mathscr{X}))^{1/2}}{N^{1/2}}.
\end{equation*}
Then, provided $N\geq 4$, the following hold: 
\begin{enumerate}
\item[(i)] (Mean estimation)
    $
    \mathbb{E}[\mathcal{W}_{1/2}(\mu,\mu^N)] 
    \leq 
    {\rm diam}(\mathscr{X})^{1/2} \min\{2\mathtt{r}_1(N), 24\mathtt{r}_2(N)\}
    ,
    $
\item[(ii)] (Concentration) with probability at least $1-\delta$
    \begin{equation*}
    \big|\mathcal{W}_{1/2}(\mu,\mu^N) - \mathbb{E}[\mathcal{W}_{1/2} (\mu,\mu^N)]\big| 
    \le 
    {\rm diam}(\mathscr{X})^{1/2}
    \min\{\mathtt{r}_1(N), 24 \mathtt{r}_2(N), \mathtt{r}_3(N)\} + \mathtt{t}(N,\delta).
    \end{equation*}
\end{enumerate}
\end{proposition}

\noindent \textit{Proof.}
See Appendix~\ref{appx:doublingembedding}.

\begin{remark} \label{rem:brief}
We make a brief remark that for $N\geq 17$, it is readily verified that $\mathtt{r}_3(N) > \mathtt{r}_1(N)$, which accounts for the absence of $\mathtt{r}_3(N)$ in the bound in 
Proposition~\ref{prop:New_Convergence__SuperAssouad}(i).
Indeed, a version derived in the provided proof takes the form
\begin{equation*}
    {\rm diam}(\mathscr{X})^{1/2}
    \min\{2\mathtt{r}_1(N), 24\mathtt{r}_2(N), 19\mathtt{r}_3(N)\}
\end{equation*}
which reduces to ${\rm diam}(\mathscr{X})^{1/2}\min\{2\mathtt{r}_1(N), 24\mathtt{r}_2(N)\}$ for all $N\in\N$.
\end{remark}

\begin{remark}
An interesting quantity in Proposition~\ref{prop:New_Convergence__SuperAssouad} is the doubling constant $\mathtt{M}$ of the $k$-point metric space $(\mathscr{X},d_{\mathscr{X}})$. 
When this metric space is the graph metric space $(G, d_{G})$ for a simple graph $G=(V,E)$ with $\operatorname{diam}(G) \leq 2$, we demonstrate in Appendix~\ref{s:AuxiliaryResults__ss:Proofs} that $\mathtt{M}$ can be explicitly bounded using familiar graph invariant.
\end{remark}

\subsection{Main technical tools for Theorem~\ref{thrm:Main_Result__random}}

The proof of Theorem~\ref{thrm:Main_Result__random} rests on two technical ingredients.
The first, given as Proposition~\ref{prop:measurability} below, concerns the measurability of $\sup_{f\in\mathcal{F}_{\rm GCN}} \mathcal{R}_{\mathbf{G},\mathbf{X}}(f)$ (given an instance of $\mathbf{G}$). This follows from a fairly direct analytic argument: one reduces the supremum over $\mathcal{F}_{\rm GCN}$ to the supremum over a suitable countable subset. While this result is presumably standard, we have not located a relevant reference in the literature and therefore provide a proof for completeness.

In what follows, we recall that the random node label is simply $\mathbf{Y}= f^{\star}(\mathbf{X})$, a random variable taking values in $E_{\rm out}^{k}$.

\begin{proposition} \label{prop:measurability} 
Let $k\in\N$. 
Let $ E_{\rm in}^{k}$ and $ E_{\rm out}^{k}$ be the feature and label spaces defined on $[k]$, respectively, where $E_{\rm in}\subset\mathbb{R}^{d_{\rm in}}$ is compact and $E_{\rm out}\subset\mathbb{R}$. 
Let $\mathcal{J}_{\mathtt{B}}$ consist of maps $f: E_{\rm in}^k\to E_{\rm out}^k$ that are $\mathtt{B}$-Lipschitz.
Then for a random feature matrix $\mathbf{X}\in E_{\rm in}^k$, the quantity 
\begin{equation*}
    \sup_{f \in \mathcal{J}_{\mathtt{B}}} \mathcal{R}_{\mathbf{X}}
    (f) \eqdef \mathbb{E}_{(V,\mathbf{Y})\sim\mathbb{P}} \big[\ell_{1/2}(\pi_V (f(\mathbf{X})),\mathbf{Y})\big].
\end{equation*}
is a well-defined random variable.
\end{proposition}

\noindent
\textit{Proof.} See Appendix~\ref{appx:measurability}.\\

Next, building on the Lipschitz regularity of $f\in\mathcal{F}_{\rm GCN}$ established in Corollary~\ref{cor:Transductive_GCNNs} (see \eqref{eq:GCNLip}) for the deterministic setting, we extend the analysis to the noisy case. Specifically, we compute the Lipschitz constant with respect to the graph metric when the input graph is fixed deterministically -- corresponding to the condition \eqref{eq:Liprequire2a} -- while allowing the input feature to be noisy. This forms the second key ingredient in the proof of Theorem~\ref{thrm:Main_Result__random}.

\begin{proposition}
\label{prop:Regularity_InducedMap}
Let $k \in \mathbb{N}$ such that $k \geq 2$. 
Let $G\in\mathcal{U}_k$ where $\mathcal{U}_k$ belongs to an admissible collection. 
For a random feature matrix $\mathbf{X}$ satisfying Assumption~\ref{ass:Features} and $f\in \mathcal{F}_{\rm GCN}$, we define the map $F_{\mathbf{X}} : [k]\to E_{\rm out}$ by $F_{\mathbf{X}}(v) \eqdef \pi_{v}(f(G,\mathbf{X}))$.
Let
\begin{equation*}
    {\rm Lip}(F_{\mathbf{X}}) \eqdef \max_{i\not=j\in [k]} 
    \frac{|F_{\mathbf{X}}(i) - F_{\mathbf{X}}(j)|}{d_G(i,j)}.
\end{equation*}
Then it holds with probability one that
\begin{equation*}
    \operatorname{Lip}(F_{\mathbf{X}}) \leq 
    2M d_{\rm in}^{1/2} \big(1 + c_k^{-1/2}(k-1)^{1/2}\big)^{tL} \prod_{l=1}^{L} \beta_l.
\end{equation*}
\end{proposition}

\noindent
\textit{Proof.} See Appendix~\ref{appx:Regularity_InducedMap}.

\section{Proof of Theorem~\ref{thrm:main_result__deterministic}} \label{sec:thm1proof}

In line with the discussion in Section~\ref{s:Prelims}, we equip $[k]\times E_{\rm out}$ with the metric $d_{[k]\times E_{\rm out}}((v,y),(v',y')) = \max\{d_G(v,v'), d_{\infty}(y,y')\}$.
Given $x \in E_{\rm in}^k$, we define {the diagonal} 
$\mathscr{D} \eqdef \{(v,\pi_v(f^{\star}(x))): v\in [k]\}$ and equip it with the metric induced by $d_{[k]\times E_{\rm out}}$. 
Denote the doubling constant of $G$ by $\mathtt{M}_G$, which satisfies $\mathtt{M}_G\geq 2$ when $k\geq 2$, and of $\mathscr{D}$ by $\mathtt{M}_{\mathscr{D}}$.
Then
\begin{equation} \label{eq:diamdiag}
    \#\mathscr{D} = k \quad\text{ and }\quad 2\leq \mathtt{M}_G \leq \mathtt{M}_{\mathscr{D}} 
    \quad\text{ and }\quad 
    {\rm diam}(\mathscr{D}) \leq {\rm diam}(G) + {\rm diam}( E_{\rm out}).
\end{equation}
For a hypothesis $f\in \mathcal{F}_{\mathtt{B}}$, we associate $\ell_{x,f}: [k]\times E_{\rm out} \to\mathbb{R}_{\geq 0}$, defined by $\ell_{x,f}(v,y) \eqdef \ell(\pi_v (f(x)), y)^{1/2}$.
Then $\ell_{x,f}|_\mathscr{D}$ is a function of $v\in [k]$ -- indeed,
\begin{equation*} 
    (\ell_{x,f}|_\mathscr{D})(v,y) = \ell_{x,f}(v,\pi_v(f^{\star}(x)))= \ell(\pi_v(f(x)),\pi_v(f^{\star}(x)))^{1/2}.
\end{equation*}
Further, by recalling \eqref{eqdef:emprisk}, \eqref{eqdef:truerisk} and that 
\begin{equation*}
    \mathbb{P} = 
    (\mathbbm{1}_{[k]}\times g_{[k]})_{\#}\mathbb{P}_{[k]} \quad\text{ and }\quad \mathbb{P}^N = 
    (\mathbbm{1}_{[k]}\times g_{[k]})_{\#}\mathbb{P}_{[k]}^N,
\end{equation*}
where $g_{[k]}(v) = \pi_v(f^{\star}(x))$, we may interpret
\begin{equation*}
    \mathcal{R}_{G,x}(f) = \mathbb{E}_{(V,Y)\sim\mathbb{P}} [\ell_{x,f}(V,Y)] \quad\text{ and }\quad \mathcal{R}_{G,x}^N(f) = \mathbb{E}_{(V,Y)\sim\mathbb{P}^N} [\ell_{x,f}(V,Y)].
\end{equation*}
Observe the following. Suppose $\ell_{x,f}|_{\mathscr{D}}$ is Lipschitz with a constant at most $2\mathtt{B}_{\ell} \mathtt{B}$, i.e.
\begin{equation} \label{eq:perculiarLip}
    |\ell(\pi_v(f(x)), \pi_v(f^{\star}(x)) - \ell(\pi_{v'}(f(x)), \pi_{v'}(f^{\star}(x))| \leq 2\mathtt{B}_{\ell} \mathtt{B} d_G(v,v').
\end{equation}
Then by invoking Kantorovich-Rubinstein duality (\citep[Remark 6.5 and Theorem 5.10(i)]{villani2009optimal}), we obtain
\begin{equation} \label{eq:RxW1/2}
    |\mathcal{R}_{G,x}(f) - \mathcal{R}^N_{G,x}(f)|
    \leq (2\mathtt{B}_{\ell} \mathtt{B})^{1/2}\mathcal{W}_{1/2}(\mathbb{P},\mathbb{P}^N).
\end{equation}
Noting \eqref{eq:diamdiag} as well as Remark~\ref{rem:brief}, we apply Proposition~\ref{prop:New_Convergence__SuperAssouad} to the metric space $(\mathscr{D}, d_{[k]\times E_{\rm out}}|_{\mathscr{D}})$ and deduce that for every $\delta\in (0,1)$,
\begin{equation}
\label{eq:prf_main__cases1}
    \mathcal{W}_{1/2}(\mathbb{P},\mathbb{P}^N) \leq ({\rm diam}(G) + {\rm diam}( E_{\rm out}))^{1/2}
    \min\{4\mathtt{r}_1(N), 48\mathtt{r}_2(N)\} + \mathtt{t}(N,\delta),
\end{equation}
with probability at least $1-\delta$, where $\mathtt{r}_i$ are given in \eqref{eq:rates} and $\mathtt{t}$ in \eqref{eq:tail}.
Substituting \eqref{eq:prf_main__cases1} into \eqref{eq:RxW1/2} and taking the supremum over $f \in \mathcal{F}_{\mathtt{B}}$ gives
\begin{multline*}
    \sup_{f\in\mathcal{F}_{\mathtt{B}}} |\mathcal{R}_{G,x}(f) - \mathcal{R}^N_{G,x}(f)|
    \\
    \leq (2\mathtt{B}_{\ell} \mathtt{B})^{1/2} \Big(({\rm diam}(G) + {\rm diam}( E_{\rm out}))^{1/2}
    \min\{4\mathtt{r}_1(N), 48\mathtt{r}_2(N)\} + \mathtt{t}(N,\delta)\Big),
\end{multline*}
with the same probability, as required for the conclusion.
Therefore, to complete the argument, it suffices to demonstrate \eqref{eq:perculiarLip}. 
However, this follows directly from the $\mathtt{B}_{\ell}$-Lipschitz continuity of the loss function $\ell: E_{\rm out} \times E_{\rm out} \to \R$ and the fact that $f, f^{\star}\in \mathcal{F}_{\mathtt{B}}$.
In particular, for any $(v,\pi_v(f^{\star}(x))), (v',\pi_{v'}(f^{\star}(x)))\in \mathscr{D}$, the following estimates hold:
\begin{align} \label{ellLipschitz1}
    \nonumber |\ell(\pi_v(f(x)),\pi_{v'}(f^{\star}(x))) - \ell(\pi_{v'}(f(x)), \pi_{v'}(f^{\star}(x)))|
    &\leq \mathtt{B}_{\ell} |\pi_v(f(x)) - \pi_{v'}(f(x))| \\
    &\leq \mathtt{B}_{\ell} \mathtt{B} d_G(v,v'),
\end{align}
and 
\begin{align} \label{ellLipschitz2}
    \nonumber |\ell(\pi_v(f(x)),\pi_{v}(f^{\star}(x))) - \ell(\pi_v(f(x)),\pi_{v'}(f^{\star}(x)))| 
    &\leq \mathtt{B}_{\ell} |\pi_v(f^{\star}(x)) - \pi_{v'}(f^{\star}(x))| \\
    &\leq \mathtt{B}_{\ell} \mathtt{B} d_G(v,v').
\end{align}
Combining \eqref{ellLipschitz1}, \eqref{ellLipschitz2}, together with the triangle inequality, we arrive at \eqref{eq:perculiarLip}. 
\qed

\section{Proof of Theorem~\ref{thrm:Main_Result__random}} \label{sec:thm2proof}

Denote $\mathcal{F}\eqdef\mathcal{F}_{\rm GCN}$ for brevity.
By Proposition~\ref{prop:measurability} that, under Assumption~\ref{ass:Features}, $\sup_{f\in\mathcal{F}} \mathcal{R}_{\mathbf{G},\mathbf{X}}(f)$ is a well-defined random variable.
We proceed to claim that, for every $\gamma>0$,
\begin{multline} \label{eq:derandomizationclaim}
    \mathbb{P}\Big(
    \sup_{f\in\mathcal{F}}
    |
    \mathcal{R}_{\mathbf{G},\mathbf{X}}(f)-\mathcal{R}^N_{\mathbf{G},\mathbf{X}}(f)
    |<\gamma
    \Big)
    \\
    \ge 
        \mathbb{P}\Big(
        \max_{G\in\mathcal{U}_k}
        \sup_{f\in\mathcal{F}}
        |
        \mathcal{R}_{G,\mathbf{X}}(f)-\mathcal{R}^N_{G,\mathbf{X}}(f)
        |<\gamma \big|
        \mathbf{G}\in\mathcal{U}_k
        \Big) 
        \mathbb{P}(\mathbf{G}\in\mathcal{U}_k).
\end{multline}
Indeed, consider the events 
\begin{align*}
    E_1 &\eqdef \big\{\max_{G\in\mathcal{U}_k}\sup_{f\in\mathcal{F}} 
    |\mathcal{R}_{G,\mathbf{X}}(f) -\mathcal{R}^N_{G,\mathbf{X}}(f)|<\gamma \text{ and } \mathbf{G}\in\mathcal{U}_k\big\} \\
    E_2 &\eqdef \big\{\sup_{f\in\mathcal{F}} 
    |\mathcal{R}_{\mathbf{G},\mathbf{X}}(f) -\mathcal{R}^N_{\mathbf{G},\mathbf{X}}(f)|<\gamma\big\}
\end{align*}
we argue that $E_1\subset E_2$. If $E_1=\emptyset$, we are done. Otherwise, take $\omega\in E_1$, which yields $\mathbf{G}(\omega)\in\mathcal{U}_k$, and more importantly
\begin{equation*}
    \sup_{f\in\mathcal{F}} 
    |\mathcal{R}_{\mathbf{G}(\omega),\mathbf{X}}(f)(\omega) -\mathcal{R}^N_{\mathbf{G}(\omega),\mathbf{X}}(f)(\omega)| \leq \max_{G\in\mathcal{U}_k}\sup_{f\in\mathcal{F}} 
    |\mathcal{R}_{G,\mathbf{X}}(f)(\omega) -\mathcal{R}^N_{G,\mathbf{X}}(f)(\omega)|<\gamma.
\end{equation*}
Thus, $\omega\in E_2$. It follows that
\begin{equation*}
    \mathbb{P}(E_2)\geq \mathbb{P}(E_1) 
    =
    \mathbb{P}\Big(
        \max_{G\in\mathcal{U}_k}
        \sup_{f\in\mathcal{F}} |
        \mathcal{R}_{G,\mathbf{X}}(f)(\omega)-\mathcal{R}^N_{G,\mathbf{X}}(f)(\omega)|
        <\gamma 
    \big| 
        \mathbf{G}\in\mathcal{U}_k
    \Big) 
    \mathbb{P}(\mathbf{G}\in\mathcal{U}_k),
\end{equation*}
which is \eqref{eq:derandomizationclaim}.
Next, recall from Assumption~\ref{ass:Features} that $\mathbf{X}$ takes values in $E_{\rm in}^k = [-M,M]^{d_{\rm in}\times k}$ with probability one. We may take $E_{\rm out}^k$ to be the maximum range of $f(\mathbf{X})$ for $f\in\mathcal{F}_{\rm GCN}$. 
Following the proof of Proposition~\ref{prop:Regularity_InducedMap}, particularly the steps \eqref{eq:diffterm}, \eqref{eq:fG_Lip_Bound}, and \eqref{eqdef:inftyprodmetric}, we deduce that
\begin{equation} \label{eq:Eoutrecall}
    {\rm diam}(E_{\rm out}^k) = {\rm diam}(E_{\rm out}) \leq \mathtt{D},
\end{equation}
where $\mathtt{D}$ is given in \eqref{eq:matD}.
Now let $\delta\in (0,1)$.
By Assumption~\ref{ass:Super-Density}(ii), for sufficiently large $k\in\N$ (particularly, for $k\geq 2$), we have $\mathbb{P}(\mathbf{G}\in\mathcal U_k)\geq 1-\delta$.
Consequently from \eqref{eq:derandomizationclaim},
\begin{align} \label{eq:chainofprobs}
    \nonumber \mathbb{P}\Big(
        \sup_{f\in\mathcal{F}}
        |
        \mathcal{R}_{\mathbf{G},\mathbf{X}}(f)-\mathcal{R}^N_{\mathbf{G},\mathbf{X}}(f)
        |<\gamma
    \Big)
    &\ge 
        \mathbb{P}\Big(
            \max_{G\in\mathcal{U}_k}
            \sup_{f\in\mathcal{F}}
            |
            \mathcal{R}_{G,\mathbf{X}}(f)-\mathcal{R}^N_{G,\mathbf{X}}(f)
            |<\gamma 
        \big|
            \mathbf{G}\in\mathcal{U}_k
        \Big) 
        \mathbb{P}(\mathbf{G}\in\mathcal{U}_k) \\
    &\geq (1-\delta)\mathbb{P}\Big(
            \max_{G\in\mathcal{U}_k}
            \sup_{f\in\mathcal{F}}
            |
            \mathcal{R}_{G,\mathbf{X}}(f)-\mathcal{R}^N_{G,\mathbf{X}}(f)
            |<\gamma 
        \big|
            \mathbf{G}\in\mathcal{U}_k
        \Big).
\end{align}
Further, suppose for $f\in\mathcal{F}$, with any fixed $G\in\mathcal{U}_k$ and $\mathbf{X}\in [-M,M]^{d_{\rm in}\times k}$, we have $f\in\mathcal{F}_{\mathtt{B}}$ for some $\mathtt{B}>0$, in the sense of \eqref{eq:Liprequire1}, \eqref{eq:Liprequire2a}. 
Then an estimate for 
\begin{equation*}
\mathbb{P}\Big(
    \max_{G\in\mathcal{U}_k}\sup_{f\in\mathcal{F}}|
    \mathcal{R}_{G,\mathbf{X}}(f)-\mathcal{R}^N_{G,\mathbf{X}}(f)|<\gamma^{\star} 
\big|
    \mathbf{G}\in\mathcal{U}_k
\Big)
\end{equation*}
with
\begin{equation} \label{eq:r}
    \gamma^{\star} = (2\mathtt{B}_{\ell} \mathtt{B})^{1/2}\Big((2 + \mathtt{D})^{1/2}
    \min\{4\mathtt{r}_1(N), 48\mathtt{r}_2(N)
    \} + \mathtt{t}(N,\delta)\Big),
\end{equation}
can be deduced from Theorem~\ref{thrm:main_result__deterministic}. Namely, we find that
\begin{equation} \label{eq:probfromdeterministicase}
    \mathbb{P}\Big(
        \max_{G\in\mathcal{U}_k}\sup_{f\in\mathcal{F}}|
        \mathcal{R}_{G,\mathbf{X}}(f)-\mathcal{R}^N_{G,\mathbf{X}}(f)|<\gamma^{\star} 
    \big|
        \mathbf{G}\in\mathcal{U}_k
    \Big) 
    \geq 
    1-\delta.
\end{equation}
Note that \eqref{eq:probfromdeterministicase} holds since, for $\mathbf{G} \in \mathcal{U}_k$, its realization lies in $\mathcal{U}_k$; combined with \eqref{eq:Eoutrecall}, this gives
\begin{equation*}
    (2\mathtt{B}_{\ell} \mathtt{B})^{1/2}\Big(({\rm diam}(G) + {\rm diam}(E_{\rm out}))^{1/2}
    \min\{4\mathtt{r}_1(N), 48\mathtt{r}_2(N)
    \} + \mathtt{t}(N,\delta)\Big) \leq \gamma^{\star}.
\end{equation*}
Thus, together, \eqref{eq:chainofprobs}, \eqref{eq:probfromdeterministicase} yield the desired conclusion:
\begin{equation*} 
    \mathbb{P}\Big(
    \sup_{f\in\mathcal{F}}
    |
    \mathcal{R}_{\mathbf{G},\mathbf{X}}(f)-\mathcal{R}^N_{\mathbf{G},\mathbf{X}}(f)
    |<\gamma^{\star}\Big) \geq (1-\delta)^2 \geq 1-2\delta.
\end{equation*}
It remains to produce and estimate $\mathtt{B}$ in \eqref{eq:r}.
To this end, we apply Corollary~\ref{cor:Transductive_GCNNs}, particularly \eqref{eq:GCNLip}, the arguments from the proof of Proposition~\ref{prop:Regularity_InducedMap} (see \eqref{eq:diffterm}, \eqref{eq:fG_Lip_Bound}), and the fact that $M\geq 1/2$, to derive an upper bound of $\mathtt{B}$ satisfying
\begin{equation*}
    d_{\rm in}^{1/2} \big(1 + c_k^{-1/2}(k-1)^{1/2}\big)^{tL} \prod_{l=1}^{L} \|W_l\|_{\rm op} \max\{2M,1\} 
    \le 
    2Md_{\rm in}^{1/2} \big(1 + c_k^{-1/2}(k-1)^{1/2}\big)^{tL} 
    \prod_{l=1}^{L} 
    \beta_l
    = \mathtt{D}.
\end{equation*}
Replacing $\mathtt{B}$ with $\mathtt{D}$, we conclude the proof.
\qed

\section*{Acknowledgements and funding}
\label{s:Acknow}
The authors would like to thank Ofer Neiman for his very helpful references on doubling constants and other pointers.  We would also like to thank Haitz S\'{a}ez de Oc\'{a}riz Borde for helpful discussions on practical considerations for transductive learning on graphs with GCNs.

A.\ Kratsios acknowledges financial support from the Natural Sciences and Engineering Research Council of Canada (NSERC) through Discovery Grant Nos.\ RGPIN-2023-04482 and DGECR-2023-00230.
A. M. Neuman acknowledges financial support from the Austrian Science Fund (FWF) under Project P~37010.
We further acknowledge that resources used in the preparation of this research were provided, in part, by the Province of Ontario, the Government of Canada through CIFAR, and the industry sponsors of the Vector Institute\footnote{\href{https://vectorinstitute.ai/partnerships/current-partners/}{https://vectorinstitute.ai/partnerships/current-partners/}}.

\appendix
\label{appendix}
\part{Appendix}

\section{Proofs of secondary results}

\subsection{Proof of Corollary~\ref{cor:Transductive_GCNNs}} \label{appx:Transductive_GCNNs}

To apply Theorem~\ref{thrm:main_result__deterministic}, it suffices to verify Lipschitz conditions \eqref{eq:Liprequire1}, \eqref{eq:Liprequire2a} for the GCN models specified in Definition~\ref{defn:GGCN}, when the graph input $G$ is fixed.
We define the linear operators
\begin{equation} \label{eq:intermediatelinear}
    \tilde{\mathfrak{L}}_l(H_l) \eqdef W_l((\Delta_G)^t H_{l-1}^{\top})^{\top} \quad\text{ for }\quad l=1,\dots,L-1,
\end{equation}
and $\tilde{\mathfrak{L}}_L(H_{L-1}) \eqdef W_L H_{L-1}$.
Their operator norms are estimated in the proposition below, and the verification of \eqref{eq:Liprequire1}, \eqref{eq:Liprequire2a} is presented subsequently.

\begin{proposition} 
\label{prop:regularity_GCNN__layer}
Let $k\in\N$ be such that $k\geq 2$.
Let $G\in\mathcal{G}_k$ such that ${\rm deg}_{-}(G)\geq 1$.
Then
\begin{equation*} 
    \|\tilde{\mathfrak{L}}_l\|_{\rm op}\leq \|W_l\|_{\rm op}\bigg(1 + \frac{ (k-1)^{1/2}}{{\rm deg}_{-}(G)^{1/2}}\bigg)^t, \quad\text{ for }\quad l=1,\dots,L-1,
\end{equation*}
and $\|\tilde{\mathfrak{L}}_L\|_{\rm op}\leq \|W_L\|_{\rm op}$.
\end{proposition}

\begin{proof}
As the second conclusion is obvious, we only prove the first.
Let $R_i \eqdef \sum_{j=1;j\neq i}^k [D_G^{-1/2}A_G D_G^{-1/2}]_{ij}$.
Then by the Cauchy-Schwarz inequality, 
\begin{align*} 
    R_i =
    \sum_{j=1;\,j\neq i}^k 
    \frac{\mathbbm{1}_{\{i\sim j\}}}{{\rm deg}(i)^{1/2} {\rm deg}(j)^{1/2}}
    &=\frac{ 1}{{\rm deg}(i)^{1/2}} 
    \sum_{j=1;\,j\neq i}^k 
    \frac{\mathbbm{1}_{\{i\sim j\}}}{{\rm deg}(j)^{1/2}} \\
    & \leq \frac{1 }{{\rm deg}(i)^{1/2}}
    \bigg(\sum_{j=1;\,j\neq i}^k \mathbbm{1}_{\{i\sim j\}} \bigg)^{1/2}
    \bigg(\sum_{j=1;\,j\neq i}^k \frac{1}{{\rm deg}(j)}
    \bigg)^{1/2} \\
    & \le 
    \bigg(\sum_{j=1;\,j\neq i}^k \frac{1}{{\rm deg}_{-}(G)}
    \bigg)^{1/2} \\
    &= \frac{(k-1)^{1/2}}{{\rm deg}_{-}(G)^{1/2}}.
\end{align*}
Further, by definition, $[\Delta_G]_{ii}=1$, and
\begin{equation*} 
    \sum_{j=1;j\not=i}^k [\Delta_G]_{ij} = \sum_{j=1;j\not=i}^k [I_k - D_G^{-1/2} A_G D_G^{-1/2}]_{ij} = R_i.
\end{equation*}
Consequently, the Gershgorin Circle Theorem \citep[Theorem 6.1.1]{horn2012matrix} implies that the eigenvalues $\{\lambda_i(\Delta_G)\}_{i=1}^k$ of $\Delta_G$ belong to the following set of discs in the complex plane $\mathbb{C}$:
\begin{align}
\label{eq:GershgorinPower}
    \nonumber 
    \{\lambda_i(\Delta_G)\}_{i=1}^k
    &\subset \bigcup_{i=1}^k 
    \bigg\{z\in \mathbb{C}: |z-[\Delta_G]_{ii}|
    \le R_i \bigg\} \\
    \nonumber &\subset \bigcup_{i=1}^k 
    \bigg\{z\in \mathbb{C}: |z-[\Delta_G]_{ii}| \le \frac{(k-1)^{1/2}}{{\rm deg}_{-}(G)^{1/2}}\bigg\} \\
    &= \bigcup_{i=1}^k
    \bigg\{z\in \mathbb{C}:\,|z-1|\le \frac{(k-1)^{1/2}}{{\rm deg}_{-}(G)^{1/2}}\bigg\}.
\end{align} 
Because $\Delta_G$ is symmetric, the spectral theorem \citep[Theorem 2.5.6]{horn2012matrix} ensures that all its eigenvalues are real.
Thus,~\eqref{eq:GershgorinPower} confines $\{\lambda_i(\Delta_G)\}_{i=1}^k$ to the interval
\begin{equation*}
    \bigg(1 - \frac{(k-1)^{1/2}}{{\rm deg}_{-}(G)^{1/2}}
    , 1 + \frac{(k-1)^{1/2}}{{\rm deg}_{-}(G)^{1/2}}\bigg),
\end{equation*}
which subsequently yields,
\begin{equation} \label{eq:spectral_boundA__complete__BEGIN}
    \|\Delta_G\|_{\rm op}=\max_{i=1,\dots,k}\, \big|\lambda_i(\Delta_G)\big|
    \le 1 + \frac{(k-1)^{1/2}}{{\rm deg}_{-}(G)^{1/2}}.
\end{equation}
It now follows from definition~\eqref{eq:intermediatelinear} and \eqref{eq:spectral_boundA__complete__BEGIN} that
\begin{equation*}
    \|\tilde{\mathfrak{L}}_l\|_{\rm op}\leq \|W_l\|_{\rm op}\|\Delta_G\|_{\rm op}^t
    \leq \|W_l\|_{\rm op}\bigg(1 + \frac{ (k-1)^{1/2}}{{\rm deg}_{-}(G)^{1/2}}\bigg)^t,
\end{equation*}
as wanted.
\end{proof}

Continuing with the proof of Corollary~\ref{cor:Transductive_GCNNs}, we immediately obtain the following from Proposition~\ref{prop:regularity_GCNN__layer},
\begin{equation*} 
    \|\tilde{\mathfrak{L}}_L\circ
    \dots \circ\tilde{\mathfrak{L}}_1 \|_{\rm op} \leq \bigg(1 + \frac{ (k-1)^{1/2}}{{\rm deg}_{-}(G)^{1/2}}\bigg)^{tL} \prod_{l=1}^{L} \|W_l\|_{\rm op}. 
\end{equation*}
Therefore, since each $\tilde{\mathfrak{L}}_l$ differs from $\mathfrak{L}_l$ at most only by a $\sigma$-activation that is $1$-Lipschitz, we deduce for $f=f(G,\cdot)$ with $f=\mathfrak{L}_L\circ\dots \circ\mathfrak{L}_1 \in\mathcal{F}_{\rm GCN}$ that
\begin{align} \label{eq:Lipnearconc2}
    \nonumber \|f(H_0) - f(H_0')\|_{\infty} &\leq \bigg(1 + \frac{ (k-1)^{1/2}}{{\rm deg}_{-}(G)^{1/2}}\bigg)^{tL} \prod_{l=1}^{L} \|W_l\|_{\rm op} \|H_0 - H_0'\|_2 \\
    &\leq d_{\rm in}^{1/2} \bigg(1 + \frac{(k-1)^{1/2}}{{\rm deg}_{-}(G)^{1/2}}\bigg)^{tL}\prod_{l=1}^{L} 
    \beta_l \|H_0 - H_0'\|_{\infty},
\end{align}
which is the condition \eqref{eq:Liprequire1}.
Next, since $E_{\rm out}$ is bounded, we have
\begin{equation} \label{eq:Lipnearconc4}
    |\pi_{v}(f(H_0)) - \pi_{v'}(f(H_0))| \leq {\rm diam}(E_{\rm out}) \leq {\rm diam}(E_{\rm out}) d_G(v,v')
\end{equation}
which is the condition \eqref{eq:Liprequire2a}. 
For the final step, we gather \eqref{eq:Lipnearconc2}, \eqref{eq:Lipnearconc4}, 
and invoke Theorem~\ref{thrm:main_result__deterministic}. The proof is now completed. \qed

\subsection{Proof of Corollary~\ref{cor:Transductive_GCNNs__random}}\label{app:proof_of_corollary2}

Corollary~\ref{cor:Transductive_GCNNs__random} follows directly from Theorem~\ref{thrm:Main_Result__random} via the next lemma. It gives an explicit bound on the typical vertex degree in a sufficiently connected Erd\H{o}s-R\'{e}nyi graph and shows that the diameter is at most $2$ with high probability. 
A qualitative version appears in~\citep{Bollobas_RGraphs_2001}, but without explicit probability estimates, which we record here for completeness.

\begin{lemma}\label{lem:Degree}
Let $\mathbf{G}=\mathbf{G}(k,p(k))$ be an Erd\H{o}s-R\'{e}nyi random graph, where $p(k)=(C\log (k)/k)^{1/2}$, with $C>2$ and $k\in\mathbb{N}$ sufficiently large.
Then the following hold:
\begin{itemize}
    \item[(i)] there exist absolute constants $c_1, c_2>0$ such that for every $\delta>0$ and $k$ large, the event
    \begin{equation} \label{eq:degreebalance}
        c_1(1- \delta ) (k\log (k))^{1/2} \leq {\rm deg}_{-}(\mathbf{G})\leq  {\rm deg}_{+}(\mathbf{G}) \leq  c_2(1+\delta ) (k\log (k))^{1/2}
    \end{equation}
    happens with probability at least $1-2k\exp \big(-(Ck\log (k))^{1/2}\delta^2 /2)\big)$;
    \item[(ii)] for sufficiently large $k$, the event
    \begin{equation*}
        \operatorname{diam}(\mathbf{G}) \leq 2
    \end{equation*}
    happens with probability at least $1-k^2\exp\big(-C (k-2)\log (k)/k\big)$.
\end{itemize}
\end{lemma}

It follows from the lemma that $\lim_{k\rightarrow \infty}\mathbb{P} (\mathbf{G}_k \in \mathcal{U}_k)=1$, with $\delta\in (0,1/2)$ and $c_k = (c_1/2) (k\log (k))^{1/2}$ in particular, which verifies Assumption~\ref{ass:Super-Density}(ii).

\begin{proof}[Proof of Lemma~\ref{lem:Degree}]
We first note that if ${\rm deg}_{+}(\mathbf{G}) \geq t$ for some $t\in\mathbb{N}$, then there must exist a vertex $v$ with ${\rm deg}(v)\geq t$.
By performing a union bound, we get
\begin{equation} \label{maxdeg}
    \mathbb{P}({\rm deg}_{+}(\mathbf{G}) \geq t)\leq k\mathbb{P}({\rm deg}(v) \geq t).
\end{equation}
Then, for any vertex $v$ and $k \geq 2$,
\begin{equation*}
    \frac{1}{2} (Ck\log (k))^{1/2}\leq \mathbb{E} [{\rm deg}(v)]= (k-1) (C\log (k)/k)^{1/2} \leq (Ck\log (k))^{1/2}.
\end{equation*}
Applying \eqref{maxdeg} and Chernoff bounds (\citep[Lemma~2.1]{chung2002connected}) for ${\rm deg}(v)$, expressed as a sum of i.i.d. Bernoulli random variables, we obtain
\begin{equation*} 
    \mathbb{P}\big({\rm deg}_{+}(\mathbf{G}) \geq (1+ \delta )(Ck\log (k))^{1/2} \big) 
    \leq k\exp \big(-(Ck\log (k))^{1/2}\delta^2 /(2+\delta )\big).
\end{equation*}
This gives the upper bound in \eqref{eq:degreebalance}. A similar argument yields the lower bound: 
\begin{equation*}
    \mathbb{P}\big({\rm deg}_{-}(\mathbf{G}) \leq (1/2)(1-\delta ) (Ck\log (k))^{1/2} \big) \leq k\exp \big(-(Ck\log (k))^{1/2}\delta^2 /2)\big),
\end{equation*}
which is the lower bound in \eqref{eq:degreebalance}.

To establish the second conclusion, we first observe that, for any two vertices, the probability that they are not adjacent and have no common neighbour is
\begin{equation*}
    (1-p(k))(1-p(k)^2)^{k-2} \leq \exp(-(k-2)p(k)^2).
\end{equation*}
Using the union bound over all $\binom{k}{2} \leq k^2/2$ vertex pairs, we deduce the probability that the diameter exceeds $2$ to be
\begin{equation*}
    \mathbb{P}(\operatorname{diam}(\mathbf{G})>2)\leq k^2 \exp\big(-(k-2)p(k)^2\big).
\end{equation*}
Substituting in $p(k)^2=C \log (k)/k$, we obtain
\begin{equation*}
    \mathbb{P}(\operatorname{diam}(\mathbf{G})>2)
    \leq 
    k^2 \exp\big(
            -(k-2)C \log (k)/k\big) 
    \leq 
    k^2 \exp\big(- C \log (k) + o(1)\big)
    =
    k^{2-C+o(1)}, 
\end{equation*}
which, since $C>2$, converges to zero as $k \rightarrow \infty$.
\end{proof}

The corollary follows from the conclusion \eqref{eq:thmrate} of Theorem~\ref{eq:thmrate} and \eqref{eq:degreebalance}, with the updated $\mathtt{D}$ given in \eqref{eq:updatedD}. \qed

\section{Supporting auxiliary results}
\label{s:AuxiliaryResults}

\subsection{Embeddings of low-distortion or of a low-dimensional representation}

Let $(\mathscr{X}, d_{\mathscr{X}}^{1/2})$ be a snowflaked version of a $k$-point doubling metric space $(\mathscr{X}, d_{\mathscr{X}})$, and let $\mathtt{M}$ denote the doubling constant of both spaces. 
We present a bi-Lipschitz embedding result, of independent interest, for $(\mathscr{X}, d_{\mathscr{X}}^{1/2})$ into $(\mathbb{R}^{m},d_{\infty})$, which also plays a key role in the proof of Proposition~\ref{prop:New_Convergence__SuperAssouad}. 

\begin{lemma}
\label{lem:independent embedding}
Let $(\mathscr{X}, d_{\mathscr{X}})$ be a $k$-point doubling metric space, with $k\geq 2$ and the doubling constant $\mathtt{M}\geq 2$. 
Assume $d_{\mathscr{X}}(x,x')\geq 1$ for all $x\not=x'\in\mathscr{X}$.
Then for the following values of $\eta\geq 0$, there exists an $m\in\mathbb{N}$ and a bi-Lipschitz embedding $\varphi_m: (\mathscr{X}, d_{\mathscr{X}}^{1/2}) \to (\mathbb{R}^m,d_{\infty})$ with distortion at most $1+\eta$, such that:
\begin{enumerate}
    \item for $\eta=0$: $m=k$,
    \item for $\eta\in (0,1/20]$: $m=\lceil 4\mathtt{M}^{5+\log_2(5)}\rceil$,
    \item for $\eta\in (1/20,1)$: $m=\lceil \eta^{-C\log_2(\mathtt{M})} \rceil$,
    \item $\eta = 12k \operatorname{diam}(\mathscr{X})^{1/2} - 1$: $m=1$.
\end{enumerate}
Here in the case $3$, $C> 1$ is an absolute constant.
In particular, when $\eta\in [1/2^{1/(C\log_2(\mathtt{M}))},1)$, we have $m=2$.
\end{lemma}

\begin{remark} 
\label{rem:trend}
A key observation from Lemma~\ref{lem:independent embedding} is that increasing the distortion allows for a reduction in the embedding dimension. Specifically, the lemma addresses either the low-distortion scenarios, where the distortion $1+\eta\in [1,2)$, or the minimal dimension case, with $m=1$. 
\end{remark}

\begin{proof}[Proof of Lemma~\ref{lem:independent embedding}]
We consider the separate cases.

\textit{Case $1$:} First, since $(\mathscr{X},d_{\mathscr{X}}^{1/2})$ is a $k$-point metric space, the Fr\'{e}chet embedding theorem \citep[Proposition 15.6.1]{MatousekLecturesBook_2002} guarantees an isometric embedding $\varphi^*:(\mathscr{X},d_{\mathscr{X}}^{1/2})\to (\mathbb{R}^{k},d_{\infty})$. 
Hence, by setting $m=k$ and $\varphi_{m}=\varphi^*$, we obtain the first conclusion.

\textit{Case $2$:} We appeal to the $\ell^{\infty}$ version of Assouad's Embedding Theorem due to~\citep[Theorem 3]{OferLowDimEmbeddingDoublingSpaces}. 
This result implies that for every $\eta\in (0,1/20]$ and every $\alpha\in (0,1)$, if we set 
\begin{equation}
\label{eq:firstrange}
    m = \bigg\lceil \frac{\mathtt{M}^{6+\log_2(1/(8\eta))}}{\alpha(1-\alpha)}\bigg\rceil = \bigg\lceil \frac{\mathtt{M}^{3+\log_2(1/\eta)}}{\alpha(1-\alpha)}\bigg\rceil,
\end{equation}
then there exists a bi-Lipschitz embedding $\varphi^*_{\eta,\alpha}:(\mathscr{X},d_{\mathscr{X}}^{1-\alpha})\to (\mathbb{R}^m,d_{\infty})$ of distortion at most $1+\eta$. 
We note that the exponent of $\mathtt{M}$ given in \eqref{eq:firstrange} can be deduced from the proof of~\citep[Theorem 3]{OferLowDimEmbeddingDoublingSpaces} together with~\citep[Proposition 2]{OferLowDimEmbeddingDoublingSpaces}.
Moreover, $m$ in \eqref{eq:firstrange} is minimized at $\alpha=1/2$ and $\eta=1/20$. Thus, we may set $m=\lceil 4\mathtt{M}^{5+\log_2(5)}\rceil$ and $\varphi_m = \varphi^*_{1/20,1/2}$. The conclusion for the case $\eta\in (0,1/20]$ now follows.

\textit{Case $3$:} 
We invoke~\citep[Theorem 6.6]{HarPeledMender_2006EmbeddingViaFastNets_SIAMJCompute} and its proof, which assures that for every $\eta\in (0,1)$\footnote{The fourth paragraph of the proof of~\citep[Theorem 6.6]{HarPeledMender_2006EmbeddingViaFastNets_SIAMJCompute} implicitly assumes that the distortion must lie in $(1,2)$.}, there exists an embedding dimension $m^*$ satisfying 
\begin{equation*}
    1 \le m^* \le \eta^{-C\log_2(\mathtt{M})},
\end{equation*}
as well as a bi-Lipschitz embedding $\varphi^*_{\eta}:(\mathscr{X},d_{\mathscr{X}}^{1/2})\to (\mathbb{R}^{m^*},d_{\infty})$, where $C\geq 1$ is an absolute constant\footnote{In fact, from the proof of \citep[Theorem 6.6]{HarPeledMender_2006EmbeddingViaFastNets_SIAMJCompute}, $C>1$.}. 
Thus, by canonically embedding $(\mathbb{R}^{m^*},d_{\infty})$ into $(\mathbb{R}^{\eta^{-C\log_2(\mathtt{M})}},d_{\infty})$ via $\iota(x_1,\dots,x_{m^*}) = (x_1,\dots,x_{m^*},0,\dots,0)$, we may, for $\eta\in (1/20,1)$, fix $m= \lceil\eta^{-C\log_2(\mathtt{M})}\rceil$ and define $\varphi_m = \iota \circ\varphi^*_{\eta}$.
Further, since $\mathtt{M}\geq 2$, 
\begin{equation*}
    \frac{1}{20} < \frac{1}{2} \leq \frac{1}{2^{1/(C\log_2(\mathtt{M}))}}.
\end{equation*}
Therefore, $[1/2^{1/(C\log_2(\mathtt{M}))},1) \subset (1/20,1)$, and for $\eta$ in this smaller range, we obtain $m = \lceil\eta^{-C\log_2(\mathtt{M})}\big\rceil = 2$ as desired.

\textit{Case $4$:}
Since $d_{\mathscr{X}}(x,x')\ge 1$ for all $x\not=x'\in \mathscr{X}$, we have
\begin{equation*}
    d_{\mathscr{X}}^{1/2}(x,x')
    \le 
    d_{\mathscr{X}}(x,x')
    \le 
    \operatorname{diam}(\mathscr{X})^{1/2}
    d_{\mathscr{X}}^{1/2}(x,x').
\end{equation*}
It follows that there exists a bi-Lipschitz map $\phi_1:(\mathscr{X},d_{\mathscr{X}}^{1/2})\to (\mathscr{X},d_{\mathscr{X}})$ with distortion at most $\operatorname{diam}(\mathscr{X})^{1/2}$. 
Next, by applying either \citep[Theorem 1]{kratsios2024tighter} or \citep[Theorem 2.1]{Matousek1990LowdimEmbeddings}, we obtain a bi-Lipschitz embedding $\phi_2:(\mathscr{X},d_{\mathscr{X}})\to (\mathbb{R},|\cdot|)$ satisfying
\begin{equation*}
    d_{\mathscr{X}}(x,x')
    \le 
    |\phi_1(x)-\phi_1(x')|
    \le
    12 k 
    d_{\mathscr{X}}(x,x').
\end{equation*}
Thus, we conclude that the composite map $\varphi_1 \eqdef \phi_2\circ \phi_1:
(\mathscr{X},d_{\mathscr{X}}^{1/2})\to (\mathbb{R},|\cdot|)$ is a bi-Lipschitz embedding with distortion at most $12k\operatorname{diam}(\mathscr{X})^{1/2}$.
\end{proof}

\subsection{A snowflake concentration result}
\label{s:SnowflakeConcentration}

We establish a variant of~\citep[Lemma 16]{hou2023instance}, adapted to the setting where $\mathbb{R}^m$ is endowed with the $\ell^{\infty}$-norm.

\begin{lemma} \label{lem:con_holder_wass}
Let $\alpha\in (0,1]$.
Let $\mathscr{X}$ be a compact subset of $\R^m$. 
Let $\mu$ be a probability measure on $\mathscr{X}$, and let $\mu^N$ be its empirical measure. 
Then for all $t > 0$ and all $N\ge 4$, 
\begin{equation*}
    \mathbb{P}\Big(\big|\mathcal{W}_{\alpha}(\mu,\mu^N) - \mathbb{E}[\mathcal{W}_{\alpha}(\mu,\mu^N)] \big| \geq t \Big)  \leq 2e^{-\frac{2Nt^2}{{\rm diam}(\mathscr{X})^{2\alpha}}},
\end{equation*}
and 
\begin{equation} \label{eq:criticaldiv}
    \mathbb{E}[\mathcal{W}_{\alpha}(\mu,\mu^N)] \leq  C_{m,\alpha} {\rm diam}(\mathscr{X}) {\rm rate}_{m,\alpha}(N)
\end{equation}
where the concentrate rate ${\rm rate}_{m,\alpha}(N)$ and the constant $C_{m,\alpha}$ are both given in Table~\ref{tab:concentration_main_table_version__FULL}. 
\end{lemma}

\begin{table}[H]
    \centering
    \begin{tabular}{@{}lll@{}}
    \toprule
    \textbf{dimension} & $\boldsymbol{{\rm rate}_{m,\alpha}}$ &  $\boldsymbol{C_{m,\alpha}}$
    \\
    \midrule
    $m<2 \alpha$ & $N^{-1/2}$ & $
    \frac{2^{m/2 - 2\alpha}}{1- 2^{m/2-\alpha}}$ \\
    $m=2 \alpha$  &  $
    \lceil \log_2(N)\rceil N^{-1/2}
    $ & $\frac1{2^{\alpha-1}\alpha}$ 
    \\
    $m>2 \alpha$  & $N^{-\alpha/m}$ & $
    2\Big(\frac{\frac{m}{2} - \alpha}{2 \alpha (1-2^{\alpha-m/2})}\Big)^{2\alpha/m}\Big(1 + \frac{\alpha}{2^{\alpha}(\frac{m}{2} - \alpha)}\Big)
    $ \\
    \bottomrule
    \end{tabular}
    \caption{Rates and constants for Lemma~\ref{lem:con_holder_wass}}
    \label{tab:concentration_main_table_version__FULL}
\end{table}

\begin{proof}[{Proof of Lemma~\ref{lem:con_holder_wass}}]
The argument closely parallels the proof of~\citep[Lemma 16]{hou2023instance}, with the focus restricted to the $\ell^{\infty}$-norm. 
In effect, this removes an extra factor of $m^{\alpha/2}$ from the expression of $C_{m,\alpha}$ given in \citep[Table 2]{hou2023instance}, consistent with the remarks on~\citep[page 414]{Kloeckner_2020Benoit}.  
We omit further details.
However, note that in the case $m=2\alpha$, the constant $C_{m,\alpha}$, without the factor $m^{\alpha/2}$, and the concentration rate ${\rm rate}_{m,\alpha}(N)$, are recorded in \citep[Table 2]{hou2023instance} as
\begin{equation} \label{oldbound}
    C_{2\alpha,\alpha} = \frac{(2\alpha)^{\alpha/2}}{\alpha 2^{\alpha+1}} 
    \quad \text{ and } \quad
    {\rm rate}_{2\alpha,\alpha}(N) = \frac{(\alpha 2^{\alpha+2} + \log_2(N))}{N^{1/2}}.
\end{equation}
Thus, to obtain a cleaner -- albeit slightly less sharp -- upper bound for the right-hand side of~\eqref{eq:criticaldiv}, we redefine
\begin{equation} \label{newbound}
    C_{2\alpha,\alpha} \eqdef \frac{1}{2^{\alpha-1}\alpha} \quad\text{ and }\quad
    {\rm rate}_{2\alpha,\alpha}(N) \eqdef \frac{\lceil\log_2(N)\rceil}{N^{1/2}}.
\end{equation}
Indeed, it can be readily verified from \eqref{oldbound}, \eqref{newbound} that when $N\geq 4$, 
\begin{equation*}
    \frac{(2\alpha)^{\alpha/2}}{\alpha 2^{\alpha+1}} \frac{(\alpha 2^{\alpha+2} + \log_2(N))}{N^{1/2}} \leq \frac{1}{2^{\alpha-1}\alpha}\frac{\lceil\log_2(N)\rceil}{N^{1/2}}.
\end{equation*}
The proof is now completed.
\end{proof}

\section{Proofs of main technical tools}
\label{s:ProofsSecondary}

\subsection{Proof of Proposition~\ref{prop:New_Convergence__SuperAssouad}} \label{appx:doublingembedding}

First, we apply Lemma~\ref{lem:independent embedding}, which states that for each
\begin{equation*}
    \eta \in (0,1/20] \cup [1/2^{1/C\log_2(\mathtt{M})},1) \cup \{12k {\rm diam}(\mathscr{X})^{1/2} -1 \}
\end{equation*}
there exist a corresponding $\tilde{D}\subset (1,2)
\cup\{12k {\rm diam}(\mathscr{X})^{1/2}\}$, an embedding dimension $\tilde{m} \in \mathbb{N}$, and a bi-Lipschitz embedding $\varphi_{\tilde{m}}: (\mathscr{X}, d_{\mathscr{X}}^{1/2}) \to (\mathbb{R}^{\tilde{m}}, d_{\infty})$, such that
\begin{equation} \label{eq:diamstretch}
    {\rm diam}(\mathscr{X})^{1/2} \leq {\rm diam}(\varphi_{\tilde{m}}(\mathscr{X})) \leq \tilde{D} {\rm diam}(\mathscr{X})^{1/2}.
\end{equation}
Here in \eqref{eq:diamstretch}, we let
\begin{equation} \label{eqdef:tildeD}
    \tilde{D} \eqdef
    \begin{cases}
        \frac{21}{20} &\text{ if } \eta\in (0,1/20]\\
        2 &\text{ if } \eta\in [1/2^{1/C\log_2(\mathtt{M})},1)\\
        12k {\rm diam}(\mathscr{X})^{1/2} & \text{ if } \eta = k\operatorname{diam}(\mathscr{X})^{1/2} - 1,
    \end{cases}
\end{equation}
and, 
\begin{equation} \label{eq:mrange}
    \tilde{m} \eqdef 
    \begin{cases}
        \lceil 4\mathtt{M}^{5+\log_2(5)} \rceil &\text{ if } \eta\in (0,1/20]\\
        2 & \text{ if } \eta\in [1/2^{1/C\log_2(\mathtt{M})},1) \\
        1 & \text{ if } \eta = 12k\operatorname{diam}(\mathscr{X})^{1/2} - 1.
    \end{cases}
\end{equation}
Observe that for our purposes, we restrict attention to three regimes: 
\begin{enumerate}
    \item the high-distortion case ($1/2^{1/C\log_2(\mathtt{M})}\le \eta<1)$,
    \item the low-distortion case ($0<\eta\le 1/20$),
    \item the extremal one-dimensional embedding case at the cost of accepting a very high distortion.
\end{enumerate}
For each fixed value of $\tilde{m}$ given in \eqref{eq:mrange}, we set 
\begin{equation*} 
    \nu\eqdef (\varphi_{\tilde{m}})_{\#}(\mu) \quad \text{ and } \quad \nu^N\eqdef (\varphi_{\tilde{m}})_{\#}\mu^N,
\end{equation*}
which are probability measures on $(\varphi_{\tilde{m}}(\mathscr{X}),d_{\infty}) \subset (\mathbb{R}^{\tilde{m}},d_{\infty})$. 
Then by invoking Lemma~\ref{lem:con_holder_wass}, for $\nu$, $\nu^N$ and $\alpha=1$, we obtain
\begin{equation} \label{eq:BenoitConcentration2}
    \mathbb{E}[\mathcal{W}_1(\nu,\nu^N)]
    \le  
    \frac{C_{\tilde{m},1} {\rm diam}(\varphi_{\tilde{m}}(\mathscr{X}))(\mathbbm{1}_{\{\tilde{m}=2\}}\log_2(N))}{N^{1/{(\tilde{m}\vee 2)}}},
\end{equation}
and for each $t>0$,
\begin{equation} \label{eq:BenoitConcentration1}
    \mathbb{P}\Big(\big|\mathcal{W}_1(\nu,\nu^N) - \mathbb{E}[\mathcal{W}_1(\nu,\nu^N)]\big|
    \geq t \Big)
    \le 
    2 e^{-\frac{2N t^2}{{\rm diam}(\varphi_{\tilde{m}}(\mathscr{X}))^2}},
\end{equation}
where the values of $C_{\tilde{m},1}$ are given in Table~\ref{tab:concentration_main_table_version__FULL}.
We translate \eqref{eq:BenoitConcentration2}, \eqref{eq:BenoitConcentration1} into expressions of Wasserstein distances between $\mu$, $\mu^N$ as follows.
By the construction given in Lemma~\ref{lem:independent embedding}, and as indicated in \eqref{eq:diamstretch}, \eqref{eqdef:tildeD}, the map $\varphi_{\tilde{m}}$ is $\tilde{D}$-Lipschitz on $(\mathscr{X}, d_{\mathscr{X}}^{1/2})$, and its inverse $\varphi_{\tilde{m}}^{-1}$ is $1$-Lipschitz on $(\varphi_{\tilde{m}}(\mathscr{X}),d_{\infty})$. 
It follows that, if $f\in \operatorname{H}(1/2,\mathscr{X},1)$ (see \eqref{stapleton}), then $f\circ \varphi_{\tilde{m}}^{-1}$ is $1$-Lipschitz on $(\varphi_{\tilde{m}}(\mathscr{X}),d_{\infty})$, and conversely, if $f\circ \varphi_{\tilde{m}}^{-1}$ is $1$-Lipschitz on $(\varphi_{\tilde{m}}(\mathscr{X}),d_{\infty})$, then $f \in {\rm H}(1/2,\mathscr{X},\tilde{D})$.
Indeed, for $x,y\in \varphi_{\tilde{m}}(\mathscr{X})$,
\begin{equation*}
    |f\circ \varphi_{\tilde{m}}^{-1}(x) - f\circ \varphi_{\tilde{m}}^{-1}(y)|
    \le 
    d_{\mathscr{X}}(\varphi_{\tilde{m}}^{-1}(x), \varphi_{\tilde{m}}^{-1}(y))^{1/2}\\
    \le \| x-y \|_{\infty},
\end{equation*}
and for $x,y\in \mathscr{X}$,
\begin{equation*}
    |f(x) - f(y)| \leq |f\circ \varphi_{\tilde{m}}^{-1}( \varphi_{\tilde{m}}(x)) - f\circ \varphi_{\tilde{m}}^{-1}(\varphi_{\tilde{m}}(y))|
    \le 
    \|\varphi_{\tilde{m}}(x) - \varphi_{\tilde{m}}(y)\|_{\infty}\\
    \le \tilde{D} d_{\mathscr{X}}(x,y)^{1/2}.
\end{equation*}
Therefore, by a change of variables, we get
\begin{equation} \label{eq:Wassineqs}
    \mathcal{W}_{1/2}(\mu,\mu^N) \le \mathcal{W}_1(\nu,\nu^N) \quad\text{ and }\quad
    \mathcal{W}_1(\nu,\nu^N) \le \tilde{D}\mathcal{W}_{1/2}(\mu,\mu^N).
\end{equation}
Now, on the one hand, combining \eqref{eq:diamstretch}, \eqref{eq:BenoitConcentration2}, \eqref{eq:Wassineqs} yields
\begin{equation} \label{preforwardexpectation}
    \mathbb{E}[\mathcal{W}_{1/2}(\mu,\mu^N)] 
    \leq 
    \frac{\tilde{D} C_{\tilde{m},1} {\rm diam}(\mathscr{X})^{1/2}(\mathbbm{1}_{\{\tilde{m}=2\}}\log_2(N))}{N^{1/(\tilde{m}\vee 2)}}.
\end{equation}
On the other hand, combining \eqref{eq:diamstretch}, \eqref{eq:BenoitConcentration1}, \eqref{eq:Wassineqs} allows us to derive, for $t\in (0,1)$, 
\begin{align} \label{eq:forwardconcentrationgen1}
    \nonumber \mathcal{W}_{1/2} (\mu,\mu^N) - \mathbb{E}[\mathcal{W}_{1/2}(\mu,\mu^N)] 
    & \le
    \mathcal{W}_1(\nu,\nu^N) -
    (1/\tilde{D}) \mathbb{E} [\mathcal{W}_1(\nu,\nu^N)]\\
    \nonumber & \le
    (1-1/\tilde{D})\mathbb{E} [\mathcal{W}_1(\nu,\nu^N)]
    + t \\
    &\le \frac{C_{\tilde{m},1} (\tilde{D}-1) \operatorname{diam}(\mathscr{X})^{1/2}(\mathbbm{1}_{\{\tilde{m}=2\}}\log_2(N))}{N^{1/(\tilde{m}\vee 2)}} + t,
\end{align}
along with, 
\begin{align} \label{eq:forwardconcentrationgen2}
    \nonumber \mathbb{E} [\mathcal{W}_{1/2} (\mu,\mu^N)] -\mathcal{W}_{1/2}(\mu,\mu^N) 
    &\le
    \mathbb{E} [\mathcal{W}_1(\nu,\nu^N)] + t/\tilde{D} - (1/\tilde{D}(\mathbb{E}[\mathcal{W}_1(\nu,\nu^N)] \\
    \nonumber &=
    t/\tilde{D} + (1-1/\tilde{D})\mathbb{E}[\mathcal{W}_1(\nu,\nu^N)] \\
    &\le
    \frac{C_{\tilde{m},1} (\tilde{D}-1) \operatorname{diam}(\mathscr{X})^{1/2}(\mathbbm{1}_{\{\tilde{m}=2\}}\log_2(N))}{N^{1/(\tilde{m}\vee 2)}} + t,
\end{align}
where we have used the fact that $\tilde{D}\in (1,2]$ in \eqref{eqdef:tildeD}.
Thus, together, \eqref{eq:forwardconcentrationgen1}, \eqref{eq:forwardconcentrationgen2} yield 
\begin{equation} \label{preforwardconcentration}
    \big|\mathcal{W}_{1/2}(\mu,\mu^N) - \mathbb{E} [\mathcal{W}_{1/2}(\mu,\mu^N)]\big| 
    \le \frac{C_{\tilde{m},1} (\tilde{D}-1) \operatorname{diam}(\mathscr{X})^{1/2}(\mathbbm{1}_{\{\tilde{m}=2\}}\log_2(N))}{N^{1/(\tilde{m}\vee 2)}} + t,
\end{equation}
which happens with probability at least $1- 2e^{-2N t^2/(\tilde{D}^2 {\rm diam}(\mathscr{X}))}$.
Substituting \eqref{eqdef:tildeD}, \eqref{eq:mrange} into \eqref{preforwardexpectation} and \eqref{preforwardconcentration} gives us a respective form of Proposition~\ref{prop:New_Convergence__SuperAssouad}(i) and (ii).
Therefore, it remains to bound $C_{\tilde{m},1}$ for the given range of $\tilde{m}$. 
From Table~\ref{tab:concentration_main_table_version__FULL}, we see $C_{1,1}=(\sqrt{2} + 1)/2<2$, $C_{2,1}=1$, and for other $\tilde{m}\geq 3$, 
\begin{equation*} 
        C_{\tilde{m},1} 
    = 
        2 \Big(
            \tfrac{\tilde{m}/2 - 1}{2(1 - 2^{1 - \tilde{m}/2})} 
        \Big)^{2/\tilde{m}}
        \Big( 1 + \tfrac{1}{\tilde{m} - 2} \Big) 
    \leq 
        4 \Big( \tfrac{\tilde{m}/2}{(2^{\tilde{m}/2} - 2)} \Big)^{2/\tilde{m}}
    \leq 
        \tfrac{4\cdot 2^{3/2}}{2^{3/2}-2} \Big( \frac{3}{2}\Big)^{\frac{2}{3}} 
    \leq 
        18,
\end{equation*}
as desired. 
\qed

\subsection{Proof of Proposition~\ref{prop:measurability}}
\label{appx:measurability}

Equip $\mathcal{J}_{\mathtt{B}}$ with the metric induced by the uniform norm. Since $E_{\rm in}^k$ is compact (see Assumption~\ref{ass:Features}), this makes $\mathcal{J}_{\mathtt{B}}$ a separable metric space.
For convenience, we use integral notation rather than expectation notation.
In this case, $\mathcal{R}_{{\mathbf{G}}, {\mathbf{X}}}(f)$ is given by (see \eqref{eqdef:truerisk})
\begin{equation*}
    \mathcal{R}_{{\mathbf{G}}, {\mathbf{X}}}(f) = \int_{\Omega} \ell_{1/2}(\pi_{{V}(\omega)}(f(\mathbf{X})),{\bf Y}(\omega)) \mathbb{P}(d\omega).
\end{equation*} 
The proof revolves around establishing that
\begin{equation} \label{eq:claim}
    \text{the map} \quad f \mapsto \mathcal{R}_{{\mathbf{G}}, {\mathbf{X}}}(f) \quad \text{is continuous on} \quad \mathcal{J}_{\mathtt{B}}.
\end{equation}
Once this holds, the separability of $\mathcal{J}_{\mathtt{B}}$ implies the existence of a countable dense subset $\mathcal{J}_{\mathtt{B}}^{\rm count}\subset\mathcal{J}_{\mathtt{B}}$ which does not depend on $\omega\in\Omega$ and such that
\begin{equation} \label{eq:countable_reduction}
    \sup_{f\in\mathcal{J}_{\mathtt{B}}^{\rm count}} \mathcal{R}_{{\mathbf{G}}, {\mathbf{X}}}(f) = \sup_{f\in \mathcal{J}_{\mathtt{B}}} \mathcal{R}_{{\mathbf{G}}, {\mathbf{X}}}(f).
\end{equation}
The left-hand side of \eqref{eq:countable_reduction} is measurable, so the right-hand side must be as well, and we have our desired conclusion. 
Subsequently, it suffices to focus on \eqref{eq:claim}.
We briefly remark that, in what follows, the argument relies only on the boundedness of $\mathbf{Y}$, which comes from the boundedness of $\mathbf{X}$ and that $\mathbf{Y}= f^{\star}(\mathbf{X})$, while $\mathbf{G}$ does not contribute.
Since the map $t\mapsto t^{\alpha}$ is concave for any $0<\alpha \le 1$, the Jensen's inequality implies that
\begin{equation} \label{eq:Jen1}
    \int_{\Omega} \ell_{1/2}(\pi_{{V}(\omega)} (f(\mathbf{X})), {\bf Y}(\omega))\mathbb{P}(d\omega) 
    \le
    \Big(\int_{\Omega} \ell(\pi_{{V}(\omega)} (f(\mathbf{X})), {\bf Y}(\omega))\mathbb{P}(d\omega) \Big)^{1/2}.
\end{equation}
Without loss of generality, we suppose that 
$\ell(0,0)=0$.
By the Lipschitz continuity \eqref{ellLip} of the loss function $\ell$,
\begin{align} \label{eq:Jen2}
    \nonumber \int_{\Omega} \ell(\pi_{{V}(\omega)} (f(\mathbf{X})), {\bf Y}(\omega))\mathbb{P}(d\omega)
    &\leq \int_{\Omega} |\ell(\pi_{{V}(\omega)} (f(\mathbf{X})), {\bf Y}(\omega)) - \ell(0,0)| \mathbb{P}(d\omega) \\
    &\leq \mathtt{B}_{\ell} \Big(\int_{\Omega} |\pi_{{V}(\omega)} (f(\mathbf{X}))| \mathbb{P}(d\omega) + \int_{\Omega} |{\bf Y}(\omega)| \mathbb{P}(d\omega) \Big)< \infty.
\end{align}
Thus, combining \eqref{eq:Jen1}, \eqref{eq:Jen2}, we obtain
\begin{equation} \label{eq:BOUNDa}
    \int_{\Omega} \ell_{1/2}(\pi_{{V}(\omega)} (f(\mathbf{X})), {\bf Y}(\omega))\mathbb{P}(d\omega) <\infty.
\end{equation}
Now let $(f_n)_{n \in \mathbb{N}} \subset \mathcal{J}_{\mathtt{B}}$ be such that $f_n \to f$ in uniform norm. Then for every $\omega\in\Omega$,
\begin{equation}
\label{eq:1}
    \ell_{1/2}(\pi_{{V}(\omega)} (f_n(\mathbf{X})), {\bf Y}(\omega)) \rightarrow
    \ell_{1/2}(\pi_{{V}(\omega)} (f(\mathbf{X})), {\bf Y}(\omega)).
\end{equation}
Using the boundedness of $\mathbf{Y}$, for each $\varepsilon>0$, let $\beta>0$ be such that 
\begin{equation} \label{eq:lambdachoice}
    \mathbb{P} (|{\bf Y}| \geq \beta) <\varepsilon \quad\text{ and }\quad 
    \int_{|{\bf Y}| \geq \beta} |{\bf Y}(\omega)| \mathbb{P}(d\omega) < \varepsilon.
\end{equation}
Let $b> \beta + \| f \|_{\infty} + 1$. It is evident that 
\begin{align} \label{eq:firstU}
    \nonumber \{\omega:|\pi_{{V}(\omega)} (f_n(\mathbf{X}))-{\bf Y}(\omega)| &> b\} \\
    \nonumber &= \bigcup_{j=1}^{k}\{\omega: {V}(\omega)=j, |\pi_{j} (f_n(\mathbf{X}))-{\bf Y}(\omega)|>b\}\\
    &= \bigcup_{j=1}^{k}\{\omega: {V}(\omega)=j, {\bf Y}(\omega) \not\in (\pi_{j} (f_n(\mathbf{X}))-b, \pi_{j} (f_n(\mathbf{X}))+b)\}.
\end{align}
Referring to \eqref{eq:1}, we henceforth consider only sufficiently large $n\in\mathbb{N}$ so that for all $j \in [k]$, the condition $\|\pi_{j} \circ f_n\|_{\infty} \leq \|f\|_{\infty}+1$ holds. In this way, we obtain from \eqref{eq:firstU}
\begin{align} \label{eq:secondU}
    \nonumber \{\omega:|\pi_{{V}(\omega)} (f_n(\mathbf{X}))-{\bf Y}(\omega)| > b\} 
    &\subset \bigcup_{j=1}^{k} \Big\{\omega:{V}(\omega)=j, {\bf Y}(\omega) \not\in \bigcap_{|z| \leq \|f\|_{\infty}+1} (z-b,z+b) \Big\}\\
    \nonumber &\subset \bigcup_{j=1}^{k}\{\omega:{V}(\omega)=j, {\bf Y}(\omega) \not\in (-\beta,\beta)\}\\
    &\subset \{\omega: |{\bf Y}(\omega)| \geq \beta\},
\end{align}
uniformly for all $n\in\mathbb{N}$ large. 
Here, in the second inclusion, we have used the fact that by the choice of $b$, we have $(-\beta,\beta) \subset (z-b,z+b)$ for all $|z| \leq \|f\|_{\infty} + 1$. 
Using \eqref{eq:secondU} along with \eqref{eq:lambdachoice}, we deduce
\begin{align} \label{eq:tailB}
    \nonumber \int_{\{\ell_{1/2} (\pi_{{V}(\omega)} (f_n(\mathbf{X})), {\bf Y}(\omega))\geq \mathtt{B}_{\ell}^{-2} b^2\}} &\ell_{1/2}(\pi_{{V}(\omega)} (f_n(\mathbf{X})), {\bf Y}(\omega)) \mathbb{P}(d\omega) \\
    \nonumber &\leq 
    \mathtt{B}_{\ell}^{1/2} \Big(\int_{\{\ell (\pi_{{V}(\omega)} (f_n(\mathbf{X})), {\bf Y}(\omega))\geq \mathtt{B}_{\ell}^{-1} b\}} |\pi_{{V}(\omega)} (f_n(\mathbf{X}))-{\bf Y}(\omega)| \mathbb{P}(d\omega) \Big)^{1/2} \\
    \nonumber &\leq
    \mathtt{B}_{\ell}^{1/2} \Big(\int_{\{|{\bf Y}| \geq \beta\}}|\pi_{{V}(\omega)}(f_n(\mathbf{X}))| \mathbb{P}(d\omega) + \int_{\{|{\bf Y}| \geq \beta\}} |{\bf Y}(\omega)| \mathbb{P}(d\omega) \Big)^{1/2} \\
    &\leq \mathtt{B}_{\ell}^{1/2} ((\|f\|_{\infty}+1)\varepsilon + \varepsilon)^{1/2},
\end{align}
expressing that $(f_n)_{n\in\N}$ is a uniformly integrable sequence.  
Applying the Vitali's convergence theorem (see \citep[Chapter~4.6]{royden2010real}) together with \eqref{eq:BOUNDa}, \eqref{eq:1}, and \eqref{eq:tailB}, we conclude that \eqref{eq:claim} holds as wanted.\qed

\subsection{Proof of Proposition~\ref{prop:Regularity_InducedMap}} \label{appx:Regularity_InducedMap}

We first show that ${\rm Lip}(F_{\mathbf{X}})$ is a random variable.
To this end, we observe the following.
Let $(E_1,d_{E_1})$ be a separable metric space, and let $(E_2,d_{E_2})$ be another metric space (not required to be separable) equipped with the Borel sigma algebra $\mathcal{B}(E_2)$.
Let $(\Omega, \mathcal{A})$ be a measurable space. 
Let $f:\Omega\times E_1\to E_2$ be such that:
\begin{enumerate}
    \item for any fixed $\omega\in\Omega$, $x\mapsto f(\omega,x)$ is Lipschitz;
    \item for any fixed $x\in E_1$, $\omega\mapsto f(\omega,x)$ is measurable.
\end{enumerate}
Then it follows, from the joint continuity of
\begin{equation*}
    (x,y)\mapsto \frac{d_{E_2}(f(\omega,x),f(\omega,y))}{d_{E_1}(x,y)},\quad x\neq y,
\end{equation*}
that the supremum of the (possibly uncountable) family of random variables $\big\{\frac{d_{E_2}(f(\omega,x),f(\omega,y))}{d_{E_1}(x,y)}\big\}_{x\neq y}$ is measurable, and that
\begin{equation} \label{eq:count_reduction}
    \sup_{x\neq y}\frac{d_{E_2}(f(\omega,x),f(\omega,y))}{d_{E_1}(x,y)} = \sup_{x\neq y,x,y\in D}\frac{d_{E_2}(f(\omega,x),f(\omega,y))}{d_{E_1}(x,y)},
\end{equation}
where $D$ is any countable dense subset of $E_1$. 
This makes the left-hand-side quantity in \eqref{eq:count_reduction} a random variable. Thus, by letting $(E_1, d_{E_1}) = ([k], d_G)$, $(E_2, d_{E_2}) = (E_{\rm out}, d_{\infty})$, we conclude that ${\rm Lip}(F_{\mathbf{X}})$ is a random variable. 
We proceed to derive an upper bound for it: 
\begin{align} \label{eq:diffterm}
    \nonumber 
    {\rm Lip}(F_{\mathbf{X}}) = \max_{i\not=j\in [k]} 
    \frac{|F_{\mathbf{X}}(i) - F_{\mathbf{X}}(j)|}{d_G(i,j)} & \le 
    \sup_{\mathbf{X}\in E_{\rm in}^k} 
    \max_{i,j\in [k]} 
    \big(|F_{\mathbf{X}}(i) - F_0(j)|
    +
    |F_0(j) - F_{\mathbf{X}}(j)| \big) \\
    \nonumber &= 
    \sup_{\mathbf{X}\in E_{\rm in}^k} 
    \max_{i,j\in [k]} 
    \big(|F_{\mathbf{X}}(i) - F_0(i)|
    +
    |F_0(j) - F_{\mathbf{X}}(j)| \big) \\
    & \le 
    \sup_{\mathbf{X}\in E_{\rm in}^k} 
    2 d_{\infty} (f(G,\mathbf{X}), 0),
\end{align}
where $F_0(i) \eqdef \pi_i(f(G,0))$, and so $F_0(i) = 0 = F_0(j)$.
Recall that for each $f\in\mathcal{F}_{\rm GCN}$, with $G\in\mathcal{U}_k$, the Lipschitz constant, in the sense of \eqref{eq:Liprequire1}, can be upper-bounded using the arguments in Corollary~\ref{cor:Transductive_GCNNs}, by at most
\begin{equation*}
    d_{\rm in}^{1/2} \bigg(1 + \frac{(k-1)^{1/2}}{{\rm deg}_{-}(G)^{1/2}}\bigg)^{tL} \prod_{l=1}^{L} \beta_l \leq d_{\rm in}^{1/2} \big(1 + c_k^{-1/2}(k-1)^{1/2}\big)^{tL} \prod_{l=1}^{L} \beta_l
\end{equation*}
which, together with \eqref{eq:diffterm}, subsequently entails 
\begin{equation} \label{eq:fG_Lip_Bound}
    {\rm Lip}(F_{\mathbf{X}}) \leq \sup_{\mathbf{X}\in E_{\rm in}^k} 2d_{\rm in}^{1/2} \big(1 + c_k^{-1/2}(k-1)^{1/2}\big)^{tL} \prod_{l=1}^{L} \beta_l 
    d_{\infty} (\mathbf{X}, 0).
\end{equation}
From Assumption~\ref{ass:Features}, we have $\sup_{\mathbf{X}\in E_{\rm in}^k} d_{\infty}(\mathbf{X},0)\leq M$ with probability one. 
Substituting this back into \eqref{eq:fG_Lip_Bound} yields the conclusion. \qed

\section{Lemmata to bound the metric doubling constant\texorpdfstring{$\mathtt{M}$}{}}
\label{s:AuxiliaryResults__ss:Proofs}

Let $G=(V,E)$ be a finite, simple (non-singleton) graph with $\operatorname{diam}(G)\leq 2$. 
We derive specific results for the doubling constant $2 \leq \mathtt{M} \leq \# V < \infty$ of the graph metric space $(G,d_G)$, showing in particular that $\mathtt{M}$ is closely related to the graph spectrum (via its adjacency matrix) and the graph degree distribution.
Details appear in Lemmas~\ref{lem:metricdoubling2},~\ref{lem:metricdoubling1} below.

\begin{lemma} 
\label{lem:metricdoubling2}
Let $G=(V,E)$ be a finite, simple (non-singleton) graph with ${\rm diam}(G)\leq 2$. 
Then it holds that $\mathtt{M} \leq {\rm deg}_{+}(G)+1$.
\end{lemma}

\begin{proof}
Let $r>0$, and let $v\in V$. 
We suppose first that ${\rm diam}(G)=1$. 
In this case, since $G$ is the complete graph on $\#V$ vertices, we observe
\begin{equation*}
    B(v,r) = 
    \begin{cases}
       B(v,r/2) = \{v\} &\text{ if } 0\leq r< 1, \\
       V &\text{ if } r \geq 1.
    \end{cases}
\end{equation*}
Consequently, $\mathtt{M} \leq \#V= {\rm deg}_{+}(G)+1$.
Suppose now that ${\rm diam}(G)=2$. 
There are four cases:
\begin{enumerate}
    \item $B(v,r)=B(v,r/2)=\{v\}$;
    \item $B(v,r)=B(v,1)$ and $B(v,r/2)=\{v\}$;
    \item $B(v,r)=B(v,2)=V$ and $B(v,r/2)=B(v,1)$;
    \item $B(v,r)=B(v,r/2)=B(v,2)=V$.
\end{enumerate}
In the first and fourth cases, $\mathtt{M}=1$, while in the second and third case, it can be checked that $\mathtt{M} \leq {\rm deg}_{+}(G)+1$. 
Combining all these cases, we arrive at the desired conclusion.    
\end{proof}

For the next result, we relate $\mathtt{M}$ to the spectral radius $\rho(G)$, the largest eigenvalue of its adjacency matrix $A_G$, and connect this back to the graph degree distribution.

\begin{lemma}
\label{lem:metricdoubling1}
Let $k,k_E\in\mathbb{N}$. 
Let $G=(V,E)$ be a finite, simple (non-singleton) graph with $k$ vertices and at most $k_E$ edges.
Suppose ${\rm diam}(G)\leq 2$.
Then
\begin{equation*} 
    \mathtt{M} 
    \le 
    \big(1 + \rho(G)\big)^4
    \le 
    8\big(1 + 2 k_{E} - (k-1){\rm deg}_{+}(G) + ({\rm deg}_{+}(G)-1){\rm deg}_{-}(G)
    \big)^2.
\end{equation*}
\end{lemma}

The proof of Lemma~\ref{lem:metricdoubling1} makes use of the relationship between $\mathtt{M}$ and its \textit{least measure doubling constant}, which we now define. 
Let $\mu\in\mathcal{P}(G)$. 
Then there exist $t_i\in [0,1]$, $i=1,\dots,\#V$, such that $\sum_{i=1}^{\#V} t_i=1$, and
\begin{equation} \label{eq:convexity}
    \mu = \sum_{i=1}^{\#V} t_i \delta_{v_i}.
\end{equation}
We say that $\mu$ is \textit{doubling}, on the graph metric space $(G,d_G)$, if there exists $0<C<\infty$ such that, for each $v\in G$ and every $r\geq 0$ 
\begin{equation} \label{eq:ratio_doubling}
    \mu(B(v,2r))
    \le C\mu(B(v,r)).
\end{equation}
We recall that $B(v,r)$ denotes a closed ball of radius $r$.
The smallest constant $C>0$ for which~\eqref{eq:ratio_doubling} holds is the measure doubling constant of $\mu$, denoted by $C_{\mu}$. 
We write $\mathcal{M}(G)$ to denote the set of doubling probability measures on $G$. 
Evidently from \eqref{eq:ratio_doubling}, $\mu\in\mathcal{M}(G)$ iff $t_i>0$ for $i=1,\dots,\#V$ in \eqref{eq:convexity}; i.e. $\mu$ has full support on $G$. 
Hence we can express 
\begin{equation} \label{Cmualt}
    C_{\mu} = 
    \sup_{v\in V, \, r\geq 0}\, \frac{\mu(B(v,2r))}{\mu(B(v,r))}.
\end{equation}
Inspired by~\citep[Definition~1.1]{SoriaTradecete_AnnFenniciMath_2019_DoublingMMS}, we define the least measure doubling constant of $G$ to be 
\begin{equation} \label{eqdef:CG}
    \mathtt{K} \eqdef \inf\{C_{\mu}: \mu\in\mathcal{M}(G) \}.
\end{equation}
The following two lemmas apply and serve as main components for the proof of Lemma~\ref{lem:metricdoubling1}, presented immediately thereafter.

\begin{lemma}
\label{lem:CGBoundDiamle2}
Let $G=(V,E)$ be a finite, simple (non-singleton) graph with ${\rm diam}(G)\leq 2$.
Then it holds that: 
\begin{itemize}
    \item[(i)] if ${\rm diam}(G)=1$ then
    \begin{equation} \label{eq:CGdiam1}
        \#V = \mathtt{K} \leq 1 + \rho(G);
    \end{equation}
    \item[(ii)] if ${\rm diam}(G)=2$ then
    \begin{equation} \label{eq:CGdiam2}
        \mathtt{K}
        \leq 1 + \rho(G).
    \end{equation} 
\end{itemize}
\end{lemma}

\begin{proof}
Let $\mu\in\mathcal{M}(G)$. 
We have that $\mu$ satisfies \eqref{eq:convexity} with $t_i>0$, $i=1,\dots,\#V$.
If ${\rm diam}(G)= 2$, then similarly to the proof of \citep[Proposition~19]{DurandCartagenaSoriaTradecete_DiscMath_2023__DoublingSpectralLinks}, we can upper bound $C_{\mu}$ with an alternative doubling constant related to $\mathtt{K}$; namely
\begin{equation} \label{relateddoublingconstant}
    C_{\mu} \leq \sup_{v\in V}\, \frac{\mu(B(v,1))}{\mu(B(v,0))}.
\end{equation}
Now by \citep[Theorem~10]{DurandCartagenaSoriaTradecete_DiscMath_2023__DoublingSpectralLinks}, 
\begin{equation} \label{spectrumrelation}
    \inf_{\mu\in\mathcal{M}(G)} \, \sup_{v\in V}\, \frac{\mu(B(v,1))}{\mu(B(v,0))} = 1+ \rho(G).
\end{equation}
Combining \eqref{relateddoublingconstant}, \eqref{spectrumrelation} with definition~\eqref{eqdef:CG}, we arrive at \eqref{eq:CGdiam2}.
If ${\rm diam}(G)=1$, then $G$ is the complete graph on $\#V$ vertices. 
In this case, for a vertex $v_i$, 
\begin{equation} \label{breakcases}
    \frac{\mu(B(v_i,2r))}{\mu(B(v_i,r))}
    =
    \begin{cases}
        1 &\text{ if } 0\leq r<1/2,\\
        \frac{1}{t_i} &\text{ if } 1/2\leq r<1,\\
        1 &\text{ if } r\geq 1.
    \end{cases}
\end{equation}
On the one hand, since $t_i\in (0,1]$ and $\inf_{i=1,\dots,\#V} t_i\leq 1/(\#V)$, we get from \eqref{Cmualt}
\begin{equation} \label{eq:Cmulowbd}
    C_{\mu} = \sup_{i=1,\dots,\#V} \frac{1}{t_i} =  \frac{1}{\inf_{i=1,\dots,\#V} t_i} \geq \#V.
\end{equation}
and consequently, $\mathtt{K}\geq \#V$. 
On the other hand, by choosing $\mu = \frac{1}{\#V}\sum_{i=1}^{\#V} \delta_{v_i}$, we obtain equality in \eqref{eq:Cmulowbd}. 
Thus, $\mathtt{K}=\#V$, which is the equality in \eqref{eq:CGdiam1}.
Moreover, \eqref{breakcases} implies for $\mu\in\mathcal{M}(G)$ that
\begin{equation*}
    \sup_{v\in V}\, \frac{\mu(B(v,1))}{\mu(B(v,0))} 
    = \sup_{v\in V}\, \frac{\mu(B(v,2r))}{\mu(B(v,r))} = C_{\mu},
\end{equation*}
which means \eqref{relateddoublingconstant} still holds. 
Consequently, in the case ${\rm diam}(G)=1$, it is automatic that $1+\rho(G)\geq \mathtt{K} =\#V$. 
\end{proof}

\begin{lemma} \label{lem:DoublingRelationship1}
Let $G=(V,E)$ be a finite, simple (non-singleton) graph with ${\rm diam}(G)\leq 2$.
Then it holds that
\begin{equation} \label{eq:KGindifferentdiams}
        \mathtt{M} 
    \leq 
        {\bf 1}_{{\rm diam}(G)=1} (\#V)^4 + {\bf 1}_{{\rm diam}(G)= 2} (1+\rho(G))^4
.
\end{equation}
\end{lemma}

\begin{proof}
Recall that $\mathtt{M}\geq 2$, since $G$ is non-singleton.
Let $\epsilon\in (0,1)$. 
By definition \eqref{eqdef:CG}, we can take $\mu\in\mathcal{M}(G)$ such that 
\begin{equation} \label{nearmin}
    C_{\mu} \leq \mathtt{K} + \epsilon.
\end{equation}
Let $r>0$, and let $v\in V$. 
By the definition of metric doubling constant, there must exist $v_1, \dots, v_{\mathtt{M}}\in V$ 
satisfying
\begin{equation*} 
    \max_{w\in B(v,r)}\,\min_{i=1,\dots,\mathtt{M}}\, d_G (w,v_i)
    \le r/2.
\end{equation*}
It follows that, for each $i=1,\dots,\mathtt{M}$
\begin{equation} \label{eq:countainement_open_closed}
    B(v_i,r/2) \subset B(v,2r).
\end{equation}
In addition, by~\citep[Chapter~15, Proposition 1.1]{LorentzAdancedProblems_BookII_1996} we can choose $v_1,\dots,v_{\mathtt{M}}$ with
\begin{equation*}
    r/2\le \min_{i,j=1,\dots,N;\,i\neq j}\, d_G(v_i,v_j).
\end{equation*}
I.e., from \eqref{eq:countainement_open_closed}, $\{v_i\}_{i=1}^{\mathtt{M}}$ is a $r/2$-{\it packing subset} of $B(v,2r)$, 
whence $B(v_i,r/4), B(v_j,r/4)$ are disjoint if $i\not=j$.
Subsequently, 
\begin{equation*}
    \mu(B(v,2r)) \geq 
    \sum_{i=1}^{\mathtt{M}}\,
    \mu(B(v_i,r/4)) \geq 
    \mathtt{M}\,
    \min_{i=1,\dots,\mathtt{M}}
    \mu(B(v_i,r/4)).
\end{equation*}
Take $i^{\star}\in \operatorname{argmin}_{i=1,\dots,\mathtt{M}} \mu (B(v_i,r/4))$.
Then, since $\mu\in\mathcal{M}(G)$,
\begin{equation} \label{doubling1}
    \mu(B(v,2r)) \geq \mathtt{M} \, \mu(B(v_{i^{\star}},r/4)) >0.
\end{equation}
Observe, $B(v,2r) \subset B(v_i, 15r/4)$, for any $i=1,\dots,\mathtt{M}$; particularly, $B(v,2r) \subset B(v_{i^{\star}}, 15r/4)$.
Combining this with \eqref{doubling1}, and substituting $r/4$ for $r$, we acquire
\begin{equation*}
    \mathtt{M} \leq \frac{\mu(B(v_{i^{\star}}, 2^{\lceil \log_2(15)\rceil} r))}{\mu(B(v_{i^{\star}},r))}
    = \frac{\mu(B(v_{i^{\star}}, 2^4 r))}{\mu(B(v_{i^{\star}},r))},
\end{equation*}
which by applying \eqref{Cmualt} repeatedly, results in
\begin{equation} \label{doubling2}
    \mathtt{M} \leq \frac{\mu(B(v_{i^{\star}}, 2^4 r))}{\mu(B(v_{i^{\star}},r))} 
    \leq C_{\mu} \, \frac{\mu(B(v_{i^{\star}},2^3 r))}{\mu(B(v_{i^{\star}}, r))} 
    \leq \dots \leq C_{\mu}^4.
\end{equation}
By integrating \eqref{doubling2} with \eqref{nearmin}, and taking the limit as $\epsilon\to 0$, we obtain $\mathtt{M} \leq \mathtt{K}^4$.
Thus far, we have not utilized the condition ${\rm diam}(G)\leq 2$. To incorporate this, we apply \eqref{eq:CGdiam1} and \eqref{eq:CGdiam2} of Lemma~\ref{lem:CGBoundDiamle2} to
\begin{equation*}
    \mathtt{M} 
    \leq {\bf 1}_{{\rm diam}(G)=1} \mathtt{K}^4 + {\bf 1}_{{\rm diam}(G)= 2} \mathtt{K}^4,
\end{equation*}
which allows us to conclude the lemma. 
\end{proof}

We now establish the proof of Lemma~\ref{lem:metricdoubling1}.

\begin{proof}[Proof of Lemma~\ref{lem:metricdoubling1}]
Since ${\rm diam}(G)\leq 2$, $G$ is connected. 
Thus,~\citep[Theorem 2.7]{DasKumar_DiscMath_2004_SpectralRadius_Bounds} applies; whence
\begin{equation*} 
    \rho(G)
    \le 
    \Big(2 k_{E} - (\#V-1){\rm deg}_{+}(G) 
    + ({\rm deg}_{+}(G)-1){\rm deg}_{-}(G)\Big)^{1/2},
\end{equation*}
which in turn implies
\allowdisplaybreaks
\begin{align}
    \nonumber
    \Big(1+ \rho(G)\Big)^{4}
    & \le 
    \Big( 1+ \Big(2 {k_{E}} - (\#V-1){\rm deg}_{+}(G) +
    ({\rm deg}_{+}(G)-1){\rm deg}_{-}(G) \Big)^{1/2}\Big)^4 \\
    \nonumber
    & \leq
    \Big( 1 + \Big(2 {k_{E}} - (\#V-1){\rm deg}_{+}(G) +
    ({\rm deg}_{+}(G)-1){\rm deg}_{-}(G)\Big)^{1/2}\Big)^4 \\
    \label{eq:specbound2}
    & \le
    8 \Big(1 + 2 {k_{E}} - (\#V-1){\rm deg}_{+}(G) + ({\rm deg}_{+}(G)-1){\rm deg}_{-}(G)\Big)^2.
\end{align}
The lemma now follows from a combination of \eqref{eq:CGdiam1}, \eqref{eq:CGdiam2}, \eqref{eq:KGindifferentdiams}, \eqref{eq:specbound2}.
\end{proof}

\bibliographystyle{plain}
\bibliography{refs,
refs_GNNGeneralization}

\end{document}

%% file: Package/ToC_Manipulation.tex
\usepackage[toc,page,header]{appendix}
\usepackage{minitoc}



%% file: Package/commenting.tex
\usepackage{marginnote}
\definecolor{MidnightBlue}{RGB}{25,25,112}
\definecolor{MidnightBlueComplementingGreen}{RGB}{25,112,25}
\definecolor{MidnightBlueComplementingPurple}{RGB}{112,25,112}
\definecolor{MidnightBlueComplementingRed}{RGB}{112,25,69}
\definecolor{coolblack}{rgb}{0.0, 0.18, 0.39}

\definecolor{deepjunglegreen}{rgb}{0.0, 0.29, 0.29}
\definecolor{applegreen}{rgb}{0.55, 0.71, 0.0}
\definecolor{WowColor}{rgb}{.75,0,.75}
\definecolor{MildlyAlarming}{rgb}{0.85,0.25,0.1}
\definecolor{SubtleColor}{rgb}{0,0,.50}
\definecolor{SubtleColor2}{rgb}{0.6,0.21,.50}

\definecolor{lasallegreen}{rgb}{0.03, 0.47, 0.19}

\definecolor{amethyst}{rgb}{0.6, 0.4, 0.8}
\definecolor{auburn}{rgb}{0.43, 0.21, 0.1}
\definecolor{americanrose}{rgb}{1.0, 0.01, 0.24}

\newcounter{margincounter}

\NewDocumentCommand{\AK}{moo}{
    \IfValueF{#2}{
                        {{\scriptsize
                            \textcolor{deepjunglegreen}{
                                \textbf{A:}
                                {#1}
                            }
                        }}
        }
    \IfValueT{#2}{\IfValueF{#3}{
                        {{\scriptsize
                            \textcolor{deepjunglegreen}{
                            \hfill\\
                                \noindent 
                                \textbf{A:}
                                \textit{{#1}}
                            \hfill\\
                            }
                        }}
        }}
    \IfValueT{#3}{
                        \marginnote{{\scriptsize
                            \textcolor{deepjunglegreen}{ 
                            \textbf{A:}
                            \textit{{#1}}
                            }
                        }}
        }
                    }

\NewDocumentCommand{\giulia}{mo}{
    \IfValueF{#2}{
                        {{\scriptsize
                            \textcolor{purple}{
                                \textbf{Giulia:}
                                \textit{{#1}}
                            }
                        }}
        }
    \IfValueT{#2}{
                        \marginnote{{\scriptsize
                            \textcolor{purple}{ 
                            \textbf{Giulia:}
                            \textit{{#1}}
                            }
                        }}
        }
                    }

\NewDocumentCommand{\nils}{mo}{
    \IfValueF{#2}{
                        {{\scriptsize
                            \textcolor{darkgreen}{
                                \textbf{Nils:}
                                \textit{{#1}}
                            }
                        }}
        }
    \IfValueT{#2}{
                        \marginnote{{\scriptsize
                            \textcolor{darkgreen}{ 
                            \textbf{Nils:}
                            \textit{{#1}}
                            }
                        }}
        }
                    }

\NewDocumentCommand{\luca}{mo}{
    \IfValueF{#2}{
                        {{\scriptsize
                            \textcolor{blue}{
                                \textbf{Luca:}
                                \textit{{#1}}
                            }
                        }}
        }
    \IfValueT{#2}{
                        \marginnote{{\scriptsize
                            \textcolor{blue}{ 
                            \textbf{Luca:}
                            \textit{{#1}}
                            }
                        }}
        }
                    }

\NewDocumentCommand{\amn}{mo}{
    \IfValueF{#2}{
                        {{\scriptsize
                            \textcolor{americanrose}{
                                \textbf{amn:}
                                \textit{{#1}}
                            }
                        }}
        }
    \IfValueT{#2}{
                        \marginnote{{\scriptsize
                            \textcolor{americanrose}{ 
                            \textbf{amn:}
                            \text{{#1}}
                            }
                        }}
        }
                    }